\documentclass[11pt]{article}

\usepackage[final]{acl}

\usepackage{times}
\usepackage{latexsym}

\usepackage[T1]{fontenc}

\usepackage[utf8]{inputenc}

\usepackage{microtype}

\usepackage{inconsolata}

\usepackage{graphicx}
\usepackage{hyperref}
\usepackage{url}
\usepackage{graphicx}
\usepackage{booktabs}
\usepackage{needspace}
\usepackage{placeins}
\usepackage{hyphenat}
\usepackage{caption}
\usepackage{tabularx}
\usepackage{makecell}
\usepackage{pifont} 
\usepackage{tabularx}
\usepackage{wrapfig}
\usepackage{multirow}
\usepackage{enumitem}
\usepackage{comment}
\usepackage[table,dvipsnames,usenames]{xcolor}
\usepackage{makecell}
\usepackage{amsmath}
\usepackage{amssymb}
\usepackage{caption} 
\usepackage{graphicx}
\usepackage{cuted} 
\usepackage{caption}
\usepackage{xspace} 
\newcommand{\ie}{\textit{i.e.},\xspace}
\newcommand{\eg}{\textit{e.g.},\xspace}

\title{Generating Attribution Reports for Manipulated Facial Images: A Dataset and Baseline}

\author{
  \textbf{Jingchun Lian}\textsuperscript{1}, 
  \textbf{Lingyu Liu}\textsuperscript{1}, 
  \textbf{Yaxiong Wang}\textsuperscript{2,$\dagger$}, 
  \textbf{Yujiao Wu}\textsuperscript{3}, \\
  \textbf{Lianwei Wu}\textsuperscript{4}, 
  \textbf{Li Zhu}\textsuperscript{1}, 
  \textbf{Zhedong Zheng}\textsuperscript{5} \\
  \\
  \textsuperscript{1}Xi'an Jiaotong University \quad
  \textsuperscript{2}Hefei University of Technology \\
  \textsuperscript{3}CSIRO \quad
  \textsuperscript{4}Northwestern Polytechnical University \quad
  \textsuperscript{5}University of Macau \\
  \\
  \texttt{15829901729@stu.xjtu.edu.cn}, \quad \texttt{wangyx@hfut.edu.cn}
}

\begin{document}
\maketitle


\begin{strip}
    \centering
    \includegraphics[width=0.95\textwidth]{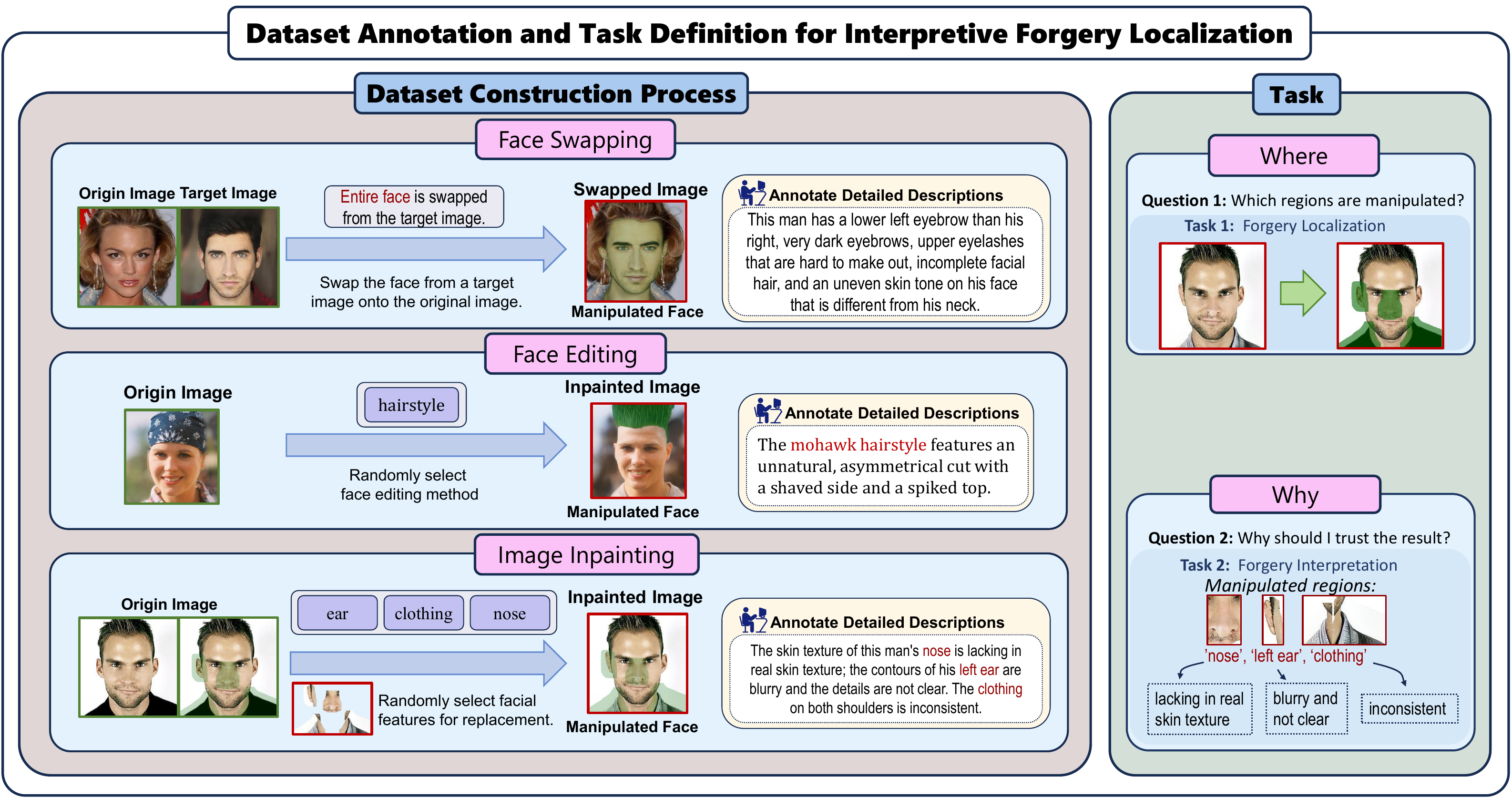}
    \captionof{figure}{An overview of our proposed benchmark, illustrating the dataset construction process and the joint task definition. The left panel shows the {3 typical} manipulation paradigms used for data generation, \ie, Face Swapping, Face Editing, and Image Inpainting. The right panel defines the task of Joint Localization and Explanation, which requires models to answer both ``where" a forgery is (Localization) and ``why" it is a forgery (Explanation).}
    \label{fig:DatasetValue}
\end{strip}

\begin{abstract}
Existing facial forgery detection methods typically focus on binary classification or pixel-level localization, providing little semantic insight into the nature of the manipulation. To address this, we introduce \textbf{Forgery Attribution Report Generation}, a new multimodal task designed to provide post-hoc forensic evidence for manipulated images. This task jointly localizes forged regions (``Where'') and generates natural language explanations grounded in the editing process (``Why''). This dual-focus approach goes beyond traditional binary forensics, providing a comprehensive, interpretable understanding of the manipulation. To enable research in this domain, we present \textbf{Multi-Modal Tamper Tracing (MMTT)}, a large-scale dataset of 152,217 samples. Each sample features a process-derived ground-truth mask and a human-authored textual description, ensuring high annotation precision and linguistic richness. We further propose \textbf{ForgeryTalker}, a unified end-to-end baseline that integrates vision and language via a shared encoder and dual decoders for mask and text generation. Experiments show that ForgeryTalker achieves competitive performance on both subtasks, i.e., 59.3 CIDEr and 73.67 IoU, establishing a strong baseline for explainable multimedia forensics. Our dataset and code are available at: \url{https://github.com/NattyLianJc/Generating-Attribution-Reports}.
\end{abstract}    
\section{Introduction}
\label{sec:intro}

The rapid evolution of advanced generative models, notably diffusion models \citep{ho2020denoising, song2020denoising,zhang2025ctrl}, has significantly enhanced the realism of synthesized images. While promising for creative domains \citep{dhariwal2021diffusion,liu2025every}, these technologies raise critical concerns regarding their misuse in misinformation and privacy violations \citep{ rana2022deepfake,liu2023deepfacelab,zhu2025seed,liu2025nablagfn,ma2024event,zeng2024multimodal}, particularly concerning facial manipulation. In response, detection research has rapidly shifted from binary classification to fine-grained forgery localization to address the growing complexity of modern attacks~\citep{verdoliva2020media, rossler2019faceforensics++,wu2023learning, yu2021survey}.

Unlike binary classifiers that merely output a simple decision, forgery localization aims to pinpoint specific tampered regions~\citep{verdoliva2020media}. However, binary masks alone provide limited interpretability~\citep{rossler2019faceforensics++}. They treat all manipulated pixels equally, failing to differentiate between subtle and significant alterations or explain the underlying rationale. Furthermore, as modern forgeries become visually indistinguishable from reality, binary masks offer insufficient guidance for human reviewers. Subtle artifacts, such as minute distortions in facial features, are often overlooked, leaving observers without descriptive evidence to verify the detected anomalies and trust the recognition results.

To address these limitations, we introduce the novel task of \textbf{Forgery Attribution Report Generation}, aiming to produce a comprehensive report consisting of a pixel-level mask and a textual explanation. To support this, we construct the \textbf{M}ulti-\textbf{M}odal \textbf{T}amper \textbf{T}racing (\textbf{MMTT}) dataset, the first large-scale benchmark with 152,217 samples. A key strength of MMTT lies in its high-quality ground truth: pixel-level masks are programmatically derived from the forgery process to ensure perfect alignment, while textual descriptions are crafted through a rigorous human-in-the-loop pipeline to capture subtle artifacts. 
Building on this, we propose \textbf{ForgeryTalker}, a unified baseline designed to generate these reports end-to-end. At its core, ForgeryTalker utilizes a shared encoder to learn a common, forgery-aware representation, forcing a deep fusion of visual and semantic features. This representation is processed by specialized dual decoders to concurrently generate the localization mask and the textual report, ensuring the explanation is semantically grounded in the visual evidence.
Notably, our framework operates under the premise that the input image has already been flagged as suspicious by an upstream binary detector~\citep{livernoche2025openfake, anan2025hybrid}. By focusing exclusively on \textbf{post-hoc attribution} (localization and explanation), we shift the forensic objective from merely delivering a binary verdict' to providing the interpretable evidence' behind it. This specialized scope ensures that computational resources are dedicated to generating meaningful semantic insights for manipulated images rather than redundant processing of pristine ones.
Our primary contributions are summarized as follows:

\begin{table*}[t]
\centering
\vspace{-.1in}
\vspace{-0.3cm}
\resizebox{\textwidth}{!}{
\begin{tabular}{lcccccc}
\toprule[1.5pt]
\textbf{Dataset}  & \textbf{Tasks} & \textbf{Modality} & \textbf{Source Samples} & \textbf{Unique Forgeries} & \textbf{Manipulation Type} & \textbf{GT Type} \\ 
\midrule
FaceForensics++~\citep{rossler2019faceforensics++}  & Class. / Seg.  & Video & 1,000 & 4,000 & Multi-Face Mods & Label + Mask \\
Celeb-DF~\citep{li2020celeb}  & Classification & Video & 590 & 5,639 & DeepFake & Image Label \\
DeeperForensics-1.0~\citep{jiang2020deeperforensics}  & Classification & Video & 50,000 & 10,000 & GAN & Image Label \\
DFDC~\citep{dolhansky2020deepfake}  & Classification & Video & 23,654 & 104,500 & DeepFake & Image Label \\
FaceShifter~\citep{li2019faceshifter}  & Classification & Video & N/A & 5,000 & GAN & Image Label \\
ForgeryNet~\citep{he2021forgerynet}  & Class. / Seg. & Image, Video & 116,321 & 221,247 & DeepFake, GAN & Label + Mask \\
OpenForensics~\citep{le2021openforensics}  & Detection / Seg. & Image, Video & 45,473 & 70,325 & GAN, Inpainting & BBox, Mask \\
DF40~\citep{yan2024df40}  & Class. / Seg. & Image, Video & N/A & $>\text{1,000,000}$ & Multi-Face Mods & Label + Mask \\ 
DiffusionFace~\citep{chen2024diffusionface} & Generation & Image & N/A & 50,000 & Diffusion & Image Label \\
GenFace~\citep{zhang2024genface}  & Generation & Image & 10,000 & 10,000 & GAN, Inpainting & Mask \\
\midrule
\rowcolor{gray!15}
\textbf{MMTT (Ours)}  & \textbf{Seg. / Caption} & \textbf{Text, Image} & \textbf{100,000} & \textbf{152,217} & \textbf{\makecell*[c]{Face Swap, Inpainting,\\ Attribute Edit}} & \textbf{Text + Mask} \\
\bottomrule[1.5pt]
\end{tabular}
}
\caption{Comparison of Face Manipulation Datasets. Our MMTT dataset is highlighted and provides rich text annotations for the \textit{``why''} problem, a unique feature among existing resources.}
\label{tab:DatasetsComparison}
\vspace{-0.4cm}
\end{table*}

\begin{itemize}[leftmargin=*]
    \item \textbf{A New Task and Dataset.} We introduce \textit{Forgery Attribution Report Generation}, a multimodal task combining forgery localization and natural language explanation. To support this, we present \textbf{M}ulti-\textbf{M}odal \textbf{T}amper \textbf{T}racing (\textbf{MMTT}), the first large-scale dataset with 152,217 samples, each annotated with a precise ground-truth mask from the editing process and a human-written attribution report.
    \item \textbf{A Unified and Effective Baseline.} We propose ForgeryTalker, {a unified framework} that jointly performs forgery localization and report generation. {It is designed to facilitate coherent cross-modal reasoning through a shared encoder (image encoder + Q-former) and dual decoders (mask decoder and a Large Language Model).}
    \item \textbf{Comprehensive Benchmarking.} We conduct extensive experiments on the proposed dataset for both report generation and forgery localization. The results validate that our baseline achieves competitive performance (59.3 CIDEr and 73.67 IoU) and demonstrates the complementary benefits of jointly addressing both tasks.
\end{itemize}

\section{{M}ulti-{M}odal {T}amper {T}racing Dataset}

\begin{figure*}[t]
    \centering
     \vspace{-0.3cm}
    \includegraphics[width=0.95\linewidth]{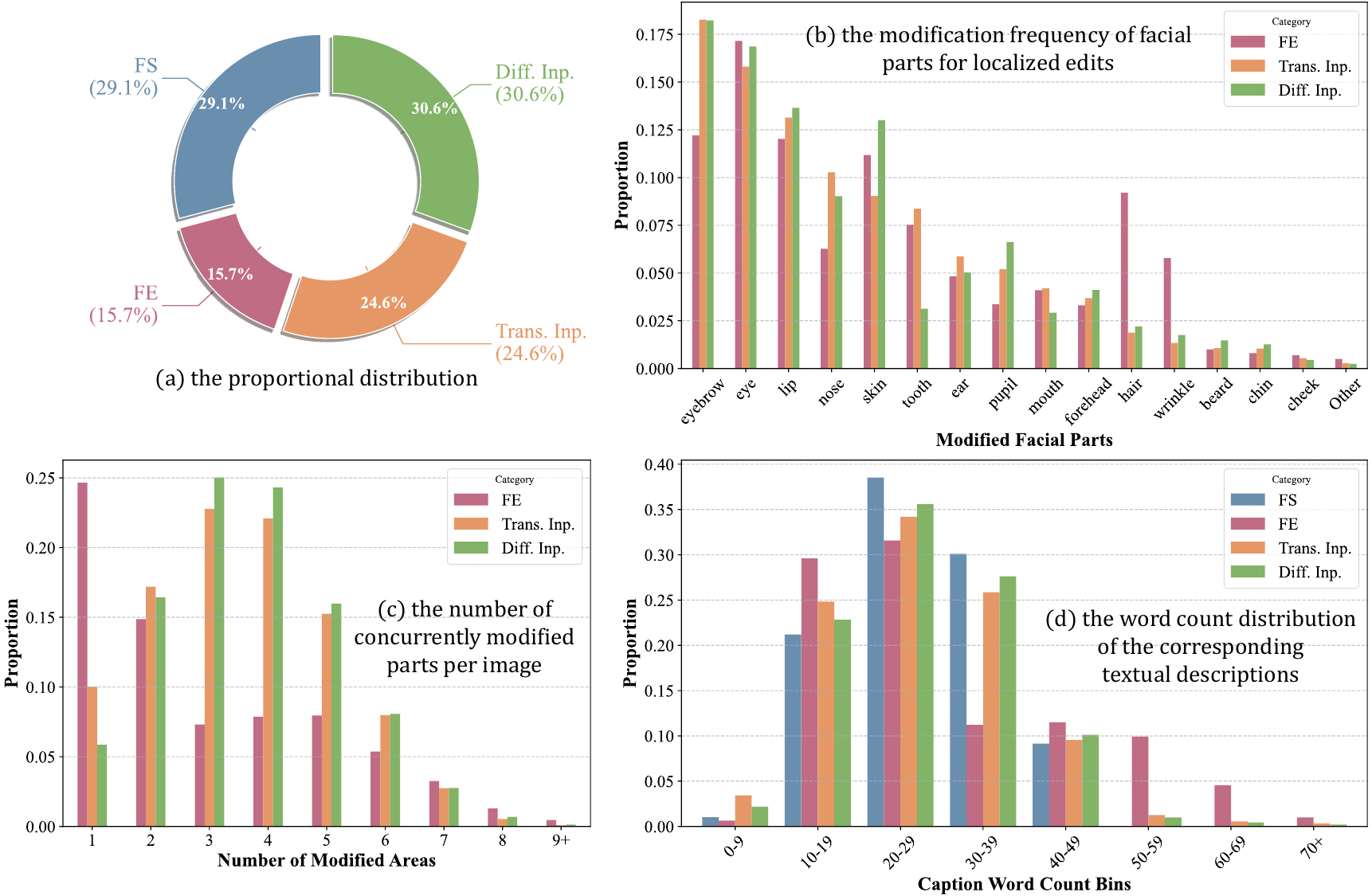}
    \vspace{-0.3cm}
    \caption{Statistical overview of the MMTT dataset and its four manipulation types: Face Swapping (FS), Face Editing (FE), Transformer-based Inpainting (Trans. Inp.), and Diffusion-based Inpainting (Diff. Inp.). The figure shows: (a) the proportional distribution of these types; (b) the modification frequency of facial parts for localized edits (excluding FS, which alters the entire face); (c) the number of concurrently modified parts per image; and (d) the word count distribution of the corresponding textual descriptions.
    }
    \label{fig:DatasetCount}
    \vspace{-0.4cm}
\end{figure*}

As a comparison with other major forgery datasets in Table~\ref{tab:DatasetsComparison} highlights, the proposed MMTT is the first to provide detailed textual explanations alongside forgery masks. The dataset contains \textbf{152,217} samples distributed across four manipulation paradigms, with diffusion-based inpainting and face swapping being the most prevalent (Figure~\ref{fig:DatasetCount}a). Our statistical analysis reveals several key properties: (1) Eyebrows, eyes, and lips are the most common targets for manipulation across all localized editing methods (Figure~\ref{fig:DatasetCount}b). (2) A significant portion of images feature multiple alterations, with 2-5 concurrent modifications being common (Figure~\ref{fig:DatasetCount}c). (3) The textual annotations form a rich corpus of over 4 million words, with an average description length of \textbf{27.4 words}. The content of these descriptions aligns closely with the visual forgeries, frequently referencing the manipulated facial parts (Figure~\ref{fig:DatasetCount}d). As shown in Figure~\ref{fig:DatasetValue}, our MMTT dataset provides two complementary types of annotations: \emph{binary forgery masks} {in Section~\ref{mask}} and \emph{forgery analysis text} {in Section~\ref{text}}. The forgery analysis text primarily delivers diagnostic summaries of facial images, while the binary masks serve as auxiliary clues, highlighting localized forgery artifacts.

\subsection{Forgery Generation}
\label{mask}
\begin{figure*}[t]
    \centering
    \vspace{-0.3cm}
    \includegraphics[width=1.0\linewidth]{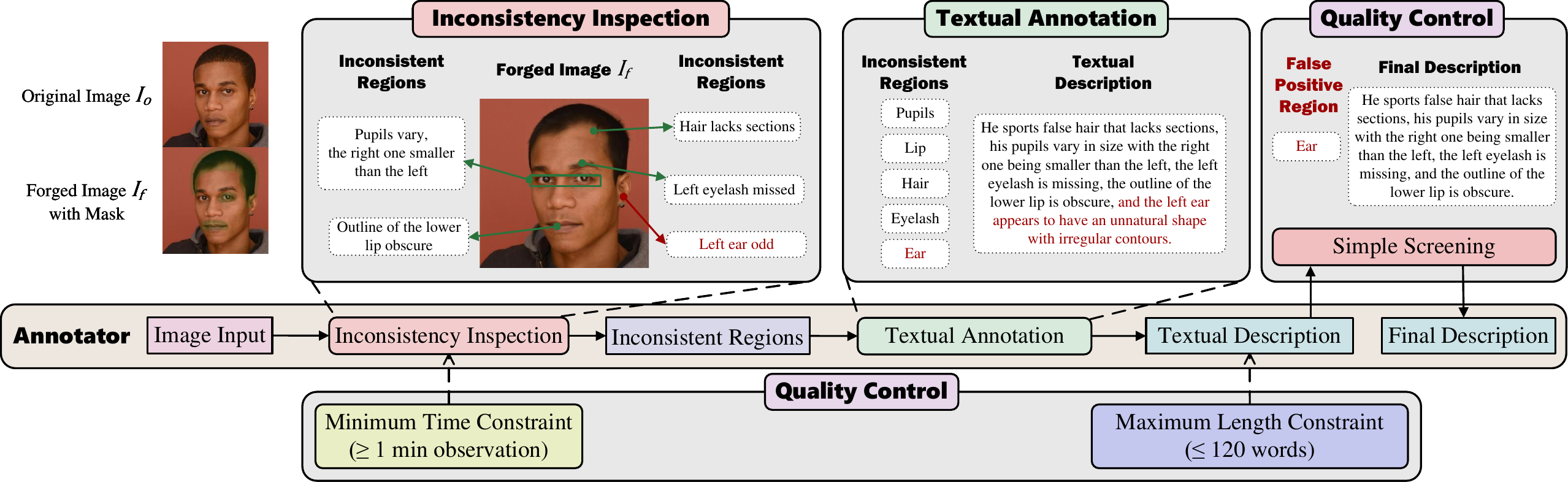}
    \vspace{-0.6cm}
    \caption{The manual annotation pipeline for the MMTT dataset. Here we show the key stages from \textcolor{pink}{inconsistency inspection} to \textcolor{OliveGreen}{textual description} and final \textcolor{magenta}{quality control}.}
    \label{fig:Annotations}
    \vspace{-0.4cm}
\end{figure*}

To construct a challenging and diverse dataset, we simulate forgery threats using three distinct manipulation paradigms. For each, we employed state-of-the-art models and developed specific procedures to programmatically generate forged images \(I_f\) and their corresponding pixel-perfect ground-truth masks \(M\).

\textbf{Source Image Collection.} We {first} construct the MMTT dataset from 100,000 high-quality facial images, comprising 30,000 images from CelebAMask-HQ~\citep{zhu2022celebv} and 70,000 images from Flickr-Faces-HQ (FFHQ)~\citep{karras2019style}. All images are resized to \(512 \times 512\) pixels, which serve as the primary source for subsequent forgery manipulations.

\noindent\textbf{Face Swapping.} We used the GAN-based E4S~\citep{e4s} model to swap faces between randomly paired images from our source datasets. Crucially, the E4S model automatically generates a precise binary mask \(M\) during this process, which directly serves as the ground-truth annotation for the manipulated region in the final forged image \(I_f\).

\noindent\textbf{Face Editing.} We performed semantic alterations using GAN-inversion models StyleCLIP~\citep{Patashnik_2021_ICCV} and HFGI~\citep{wang2021HFGI}. The transformation is applied to an input image \(I\) to produce the forged image \(I_f = \mathcal{E}_{\text{model}}(I, a)\), where \(\mathcal{E}\) is the editing function guided by attribute \(a\), and \(\text{model} \in \{\text{StyleCLIP}, \text{HFGI}\}\). The corresponding ground-truth mask, \(M_{\text{final}}\), is constructed by taking the union of any pre-existing mask (\(M_{\text{prev}}\)) and a new semantic mask (\(M_{\text{semantic}}\)) generated via a face parsing model~\citep{yu2018bisenet}, formulated as \(M_{\text{final}} = M_{\text{prev}} \cup M_{\text{semantic}}\).

\noindent\textbf{Image Inpainting.} We generated localized forgeries using both transformer-based (MAT~\citep{mat}) and diffusion-based (SDXL~\citep{sdxl}) models. The required input masks (\(M\)) were created by programmatically selecting and merging facial component segments. The final inpainted image \(I_f\) is produced by composing the original image \(I\) with the model's output \(I_g^{\text{model}}\) using the mask \(M\): $I_f = (1 - M) \cdot I + M \cdot I_g^{\text{model}}$, where \(\text{model} \in \{\text{MAT}, \text{SDXL}\}\).

\begin{figure*}[t]
    \centering
     \vspace{-.2in}
    \includegraphics[width=1\linewidth]{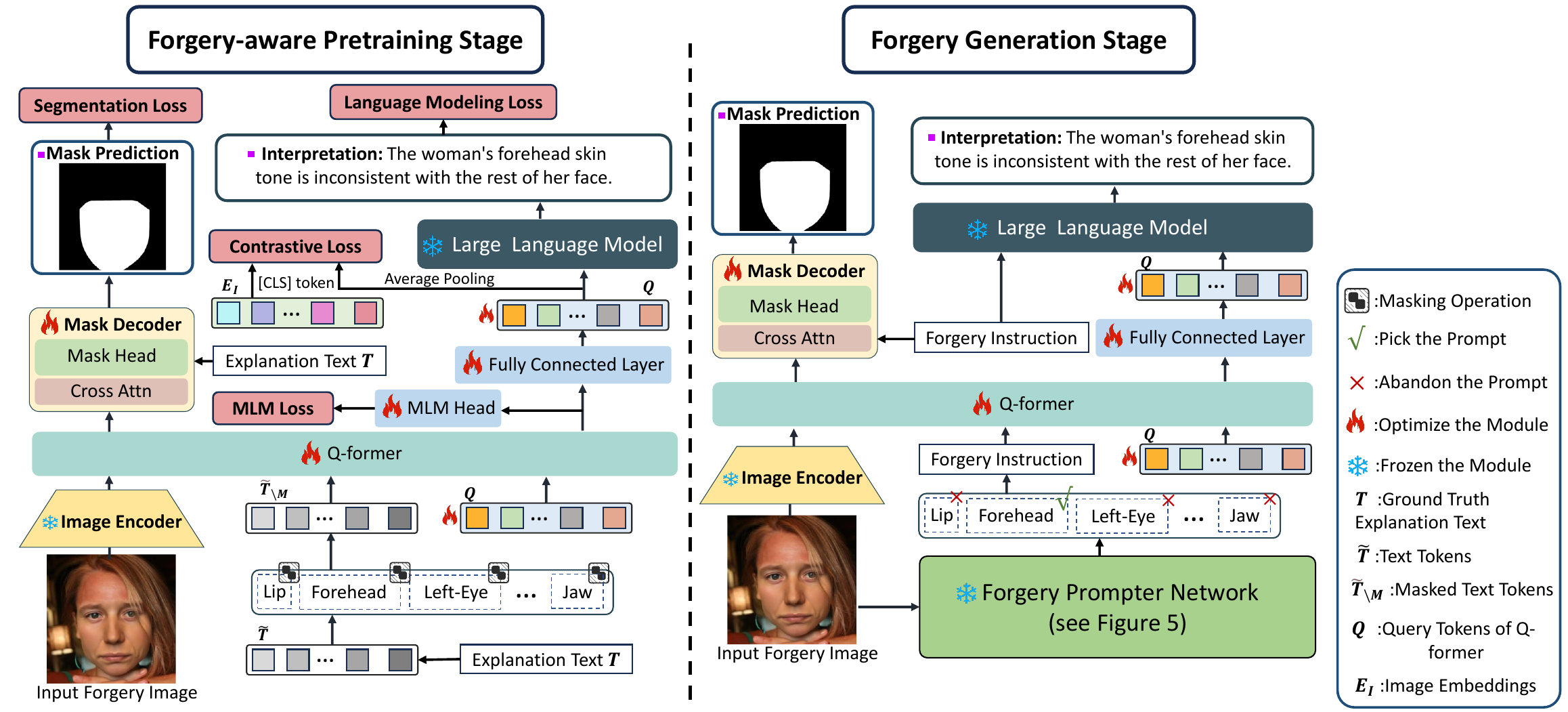}
    \vspace{-.2in}
    \caption{Illustration of our ForgeryTalker framework. The training pipeline has two stages. In the Forgery-aware Pretraining Stage, the Q-former, Mask Decoder, and Language Model are jointly optimized with MLM, language modeling, segmentation, and contrastive losses to build multimodal representations. In the Explanation Generation Stage, the FPN is trained with BCE and Dice losses for region classification and then frozen while the Q-former and Mask Decoder are fine-tuned for improved forgery localization and explanation. Finally, the multimodal features are fed to a Large Language Model to generate explanatory reports.}
    \label{fig:architecture}
    \vspace{-0.3cm}
\end{figure*}

\subsection{Diagnosis Text Annotation}
\label{text}
\noindent\textbf{Annotation Methodology.}
To ensure high-quality annotations, we develop a structured pipeline (see Figure~\ref{fig:Annotations}) where a team of expert annotators receives specific guidelines. Guided by a ground-truth mask for each image pair, annotators are instructed to describe visual inconsistencies or artifacts exclusively within the manipulated regions. They focus only on unnatural or poorly-integrated features, omitting descriptions of authentic areas, and compose self-contained descriptions that do not reference the original image. Each description is limited to a maximum of 120 words.

\noindent\textbf{Annotation Process.}
Our annotation process involves 30 trained annotators who follow a three-step procedure. First, annotators receive an original-forgery image pair $(I_o, I_f)$ with its corresponding ground-truth mask $M$. They then inspect the images for inconsistencies within the masked facial regions, such as unnatural textures or asymmetries. Finally, they compose a detailed textual description $T$, explaining the specific nature of the alteration. This process culminates in a triplet $p = (I_f, M, T)$.


\noindent\textbf{Annotation Quality Assurance.}
To ensure description reliability and inter-annotator consistency, we implement a rigorous human-in-the-loop quality control pipeline. All 30 annotators were recruited from a professional data annotation agency and provided with standardized guidelines and reference examples to maintain a uniform description style. During the process, we enforce a minimum observation time of one minute per image pair to ensure thorough examination. Furthermore, textual descriptions referencing regions outside the ground-truth mask are automatically flagged for review to prevent false positives. Finally, manual audits and cross-verifications were conducted on sampled batches to ensure that all textual descriptions are highly consistent and strictly grounded in the localized manipulation masks.

\section{ForgeryTalker}
\label{Method}




The architecture of our baseline model, \textbf{ForgeryTalker}, extends InstructBlip~\citep{dai2023instructblip} and is structured around a shared encoder and dual decoders. The shared encoder, consisting of a Vision Transformer and a Q-Former, processes the tampered image \(I\) to extract multimodal features. Guided by prompts from an integrated Forgery Prompter Network (FPN), these features are then passed to two decoders: a Mask Decoder for forgery localization and a Large Language Model (LLM) that generates the final attribution report. As shown in Figure~\ref{fig:architecture}, training proceeds in two stages. In the \textbf{Forgery-aware Pretraining Stage}, we jointly optimize the core modules using a weighted combination of losses to build forgery-sensitive multimodal representations. In the subsequent \textbf{Attribution Report Generation Stage}, we first train the FPN to generate accurate region prompts. Then, with the FPN fixed, we fine-tune the mask decoder and Q-Former using segmentation and language modeling losses to improve forgery localization and the final attribution report.

\subsection{Forgery-aware Pretraining}

To learn robust multimodal representations sensitive to manipulation artifacts, we jointly optimize the proposed model using paired image \(I\) and text \(T\) via four complementary objectives:

\noindent \textbf{Masked Language Modeling (\(\mathcal{L}_{mlm}\)).} To enforce local visual-linguistic alignment, we mask a subset of region-related tokens in \(T\) (e.g., facial parts). The Q-Former is tasked with reconstructing these masked tokens conditioned on the image embeddings and learnable queries.

\noindent \textbf{Language Modeling (\(\mathcal{L}_{lm}\)).} We employ a standard generation objective where the T5-based decoder autoregressively generates the explanation text \(\hat{T}\). This is supervised by maximizing the likelihood of the ground-truth tokens given the visual context.

\noindent \textbf{Forgery Localization (\(\mathcal{L}_{seg}\)).} To capture pixel-level anomalies, the mask decoder predicts a dense binary forgery map from the fused multimodal features. We apply a pixel-wise cross-entropy loss to align the prediction with the corresponding ground-truth manipulation mask.

\noindent \textbf{Cross-model Alignment (\(\mathcal{L}_{con}\)).} We further align the global image token and the mean-pooled text feature via contrastive learning. This objective pulls paired visual-text representations closer in the latent space while pushing unpaired ones apart.

\noindent The final overall objective is a weighted sum of these four complementary losses, equipping the model with both fine-grained localization capabilities and global semantic context.

\subsection{Forgery Prompter Network}



\begin{figure}
    \centering
    \vspace{-0.2cm}
    \includegraphics[width=\linewidth]{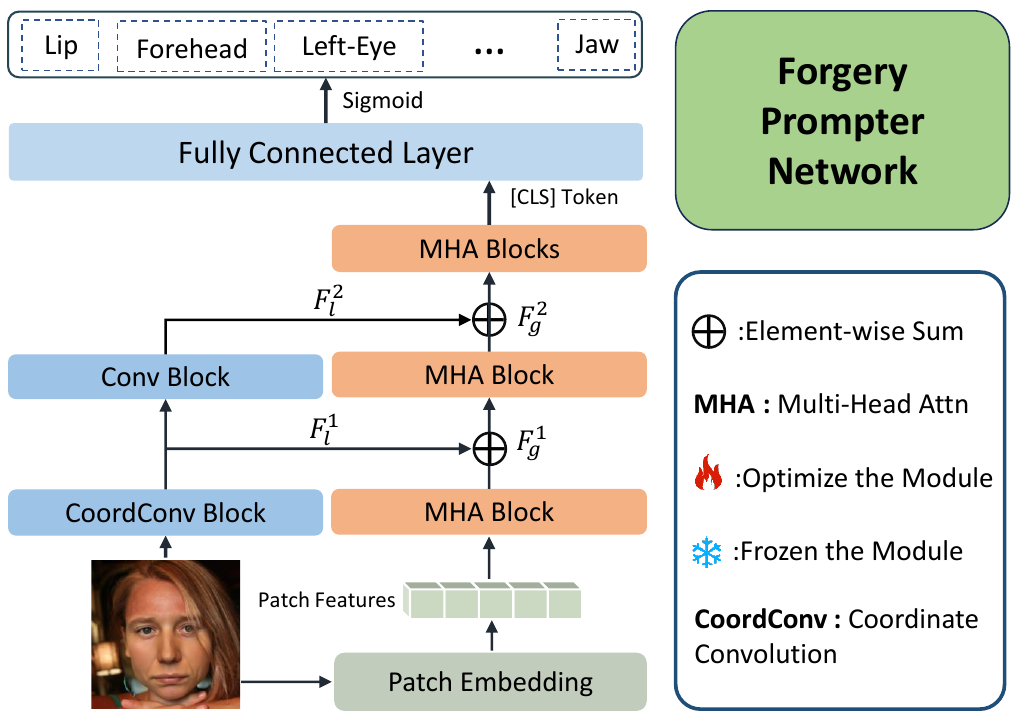}
    \caption{Illustration of the Forgery Prompter Network (FPN). The FPN generates region-aware prompts for forgery localization.}
    \label{fig:FPN}
    \vspace{-.2in}
\end{figure}

\textbf{Motivation.} 
Accurately identifying the most salient manipulated regions in forged images is challenging due to the high visual fidelity of modern manipulation techniques. Even human reviewers must closely inspect the images to detect inconsistencies. Therefore, we propose the Forgery Prompter Network (FPN) to generate an initial set of salient region keywords, which guide downstream reasoning and facilitate the coherent generation of {attribution reports}.

\noindent\textbf{Region Keywords Extraction.}
In this step, we extract relevant region labels from the provided {textual descriptions}. The defined label space comprises 21 distinct facial semantics, where each image is associated with a 21-dimensional ground-truth vector $Y$; here, the $i$-th element is 1 if the corresponding facial part is mentioned in the {textual description}, and 0 otherwise.

\noindent\textbf{Forgery Prompter
Network (FPN)} takes the vision transformers as the main architecture. Considering the crucial role of fine-grained local context in identifying subtle flaws, we introduce a convolution branch at the early $m$ layers to complement the global contexts captured by the vision transformer. As shown in Figure~\ref{fig:FPN}, the forgery image $I$ concurrently traverses self-attention blocks and convolution blocks in parallel, producing global-aware features $F_g=\{F_g^0, F_g^2, ..., F_g^{m-1}\}$ and local-aware features $F_l=\{F_l^0, F_l^2, ..., F_l^{m-1}\}$. At each encoding level, the corresponding features are element-wise summed and fed into next attention block:
 \begin{align}
     F_g^i &= \text{MHA}_{i-1}(F_g^{i-1}), F_l^i =\text{Conv}_{i-1}(F_l^{i-1}), \\
     F_g^i &= \text{MHA}_i(F_g^{i}+F_l^i), \quad i=1,\cdots,m
\end{align}
where ``MHA'' and ``Conv'' mean the multi-head attention and convolution. Furthermore, we note that the positioning of facial regions in a natural image follows a rigid and predictable structure, with the eyes typically positioned laterally relative to the nose and the eyebrows aligned above the eyes.
Leveraging this regularity, we integrate coordinate convolution \citep{liu2018intriguing} in the initial convolutional layer to detect anomalies in the arrangement of facial features, \emph{i.e.,} $\text{Conv}_{0}$ = CoorConv. 
The resultant feature $F_g^m$ contains both global and local contexts and is fed into subsequent multi-head attention blocks and a classification head to produce the probability $\hat{Y}$ across regions, while also being used in cross-attention with Q-former features to enhance forgery localization. Finally, the forgery prompter network is trained using a combined loss, incorporating both Binary Cross-Entropy (BCE) and Dice loss to effectively balance region classification and overlap precision:
\begin{equation}
\label{bceloss}
    \mathcal{L}_{BCE} = -\frac{1}{21}\sum_{i=1}^{21} Y_i\log \hat{Y}_i + \omega (1-Y_i)\log (1-\hat{Y}_i),
\end{equation}
where $\omega$ is a discount factor set to $\omega < 1$ to address the imbalance due to the prevalence of unmodified regions. The Dice loss is employed to measure the overlap between the predicted labels $\hat{Y}$ and ground truth $Y$, ensuring that less frequent classes receive more attention:
\begin{equation}
\label{diceloss}
    \mathcal{L}_{Dice} = 1 - \frac{2\sum_{i=1}^{21} Y_i \hat{Y}_i}{\sum_{i=1}^{21} Y_i + \sum_{i=1}^{21} \hat{Y}_i}.
\end{equation}
{Finally, we optimize FPN with} the average of the BCE and Dice losses via $\frac{1}{2}(\mathcal{L}_{BCE} + \mathcal{L}_{Dice})$.

\subsection{Attribution Report Generation}

Subsequently, we fix the trained FPN network and take its region predictions as prior clues to aid both the report generation and the cross-attention process for improved forgery localization. Assume the set of regions from the FPN is $R=\{r_1, r_2, ...\}$. We design a particular template to include $R$ and form a report-focused instruction $\texttt{T}_{instr}$:

\texttt{These facial areas may be manipulated by AI: [R]. Please describe the specific issues in these areas.}

This structured prompt serves as the guiding context for the language model, thereby ensuring that the final output accurately reflects the manipulations detected by the FPN. This integration enhances the coherence and quality of the generated {reports}, offering a comprehensive understanding of the tampered regions. Subsequently, the instruction and the image embeddings are fed into the Q-former, and the resulting features are passed to the large language model to generate the {explanatory text} $\hat{T}$ {with the length of \(L_{\hat{T}}\)}. This output is then supervised by the language modeling loss as:
\begin{equation}
\label{eq:lm_loss_dataset}
    \footnotesize
    \mathcal{L}_{t} = -\mathbb{E}_{(I,T)\sim\mathcal{D}}\left[\sum_{k=1}^{L_{\hat{T}}} \log P\Big(\hat{T}_k \mid I, \hat{T}_0, \ldots, \hat{T}_{k-1}\Big)\right],
\end{equation}
where \((I,T)\sim\mathcal{D}\) indicates that the expectation is taken over samples from the dataset \(\mathcal{D}\). 

\begin{table*}[t!]
    \begin{center}
        \resizebox{1\textwidth}{!}{%
            \begin{tabular}{clcccccccccc}
            \toprule[1.5pt]
                \multirow{2}{*}{\textbf{Setting}} & \multirow{2}{*}{\textbf{Method}} & \multirow{2}{*}{\textbf{Reference}} & \multicolumn{6}{c}{\textbf{Report Generation}} & \multicolumn{3}{c}{\textbf{Forgery Localization}} \\
                \cmidrule(r){4-9} \cmidrule(r){10-12}
                & & & {CIDEr} & {BLEU-1} & {BLEU-2} & {BLEU-3} & {BLEU-4} & {ROUGE-L} & {IoU}  & Precision & Recall\\ 
                \hline
                \multirow{4}{*}{\rotatebox{90}{\textbf{Zero-Shot}}} 
                & Seed1.5VL~\citep{guo2025seed15vltechnicalreport} & arXiv25 & 0.54 & 17.19 & 5.35 & 1.86 & 0.98 & 13.74 & - & - & - \\
                & Qwen2.5VL (72B)~\citep{bai2025qwen25vltechnicalreport} & arXiv25 & 2.72 & 20.34 & 6.33 & 2.55 & 1.46 & 16.52 & - & - & - \\
                & llava (72B)~\citep{liu2024llavanext} & Blog24 & 3.06 & 22.01 & 8.35 & 3.30 & 1.80 & 16.83 & - & - & - \\
                & InternVL3 (78B)~\citep{zhu2025internvl3exploringadvancedtraining} & arXiv25 & 2.67 & 22.31 & 8.08 & 3.16 & 1.75 & 16.82 & - & - & - \\
                \hline
                \multirow{8}{*}{\rotatebox{90}{\textbf{Fine-Tuning}}} 
                & PSCC-NET~\citep{liu2022pscc} & CVPR22 & - & - & - & - & - & - & 32.33 & 70.30 & 37.44 \\
                & SCA~\citep{huang2024segment}    & CVPR24 & 40.6  & 30.6  & 17.8 & 11.2 & 8.2  & 27.6 & 46.69 & 48.49 & \textbf{92.11} \\ 
                & LISA-7B~\citep{lai2024lisa}  & ICCV23 & 44.1  & 31.1  & 17.9 & 10.8 & 8.5  & 28.4 & 52.45 & 73.26 & 71.53 \\
                & Osprey~\citep{yuan2024osprey} & CVPR24 & 24.5 & 28.7 & 16.4 & 9.4 & 6.2 & 25.9 & - & - & -\\ 
                & InstructBLIP~\citep{dai2023instructblip} & NeurIPS23 & 51.7    & 31.8  & 20.3  & 14.6   & 11.4   & 27.7 & 64.04 & 87.88 & 78.53\\ 
                & FFAA~\citep{huang2024ffaa} & arXiv24 & 17.4    & 12.0  & 6.6  & 4.0   & 12.9   & 21.7 & - & - & -\\ 
                & FakeShield~\citep{xu2024fakeshield} & ICLR25 & 10.0    & 9.1  & 4.3  & 2.3   & 12.3   & 16.7 & 47.44 & 58.42 & 66.37\\ 
                \cline{2-12}
                & \textbf{ForgeryTalker} & - & \textbf{59.3}   & \textbf{35.0}  & \textbf{22.1}  & \textbf{16.0}   & \textbf{12.5}   & \textbf{28.8} & \textbf{73.67} & \textbf{91.43} & 86.22\\
            \bottomrule[1.5pt]
            \end{tabular}
        }
    \end{center}
    \caption{Performance comparison of generated captions and forgery localization across models. The "Zero-Shot" section evaluates large vision-language models without task-specific training, while "Fine-Tuning" includes specialized forgery detection and localization methods.}
    \label{tab:comparison}
\end{table*}

\begin{figure*}[t!]
    \centering
    \includegraphics[width=1.0\linewidth]{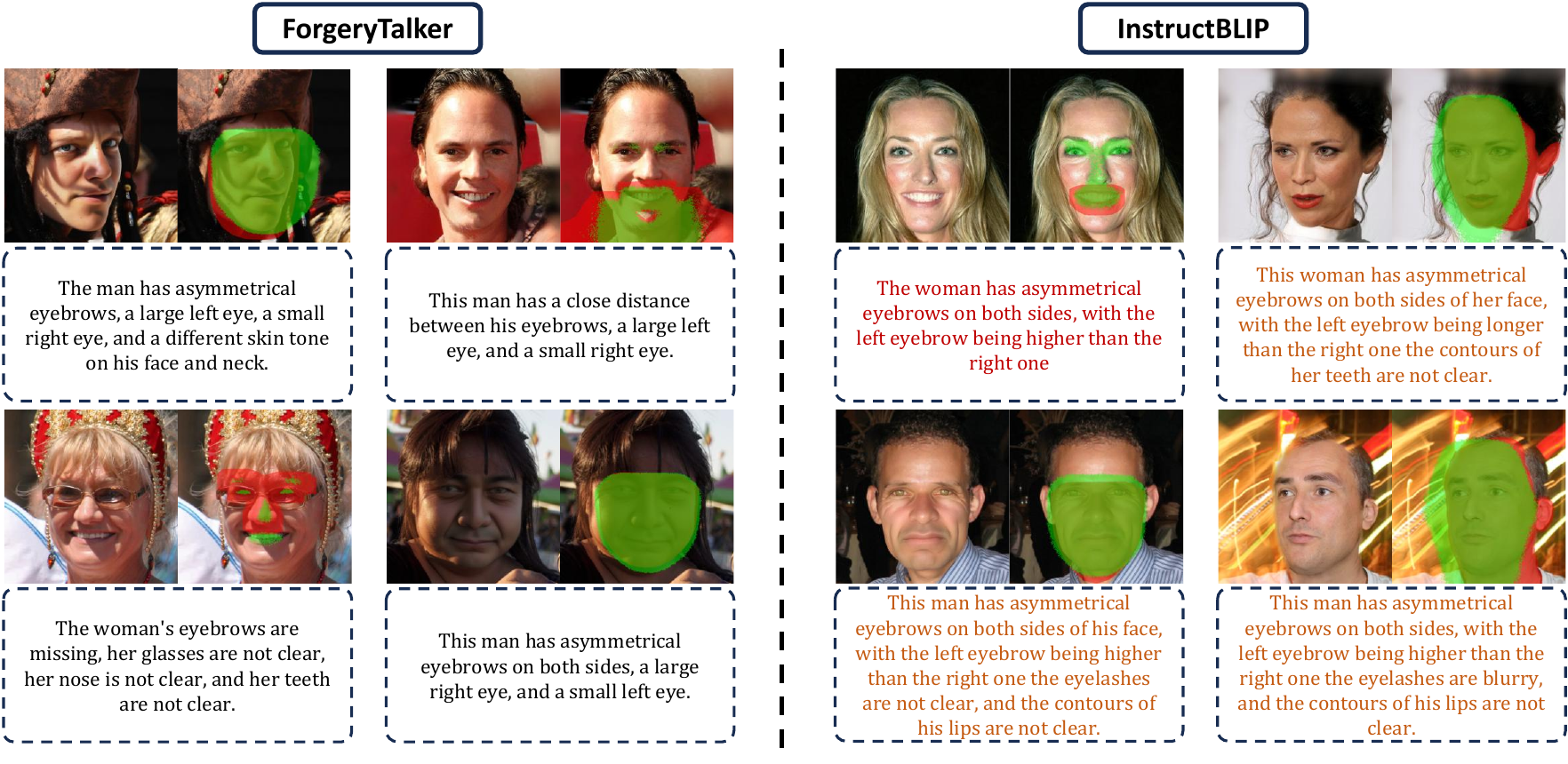}
    \captionsetup{aboveskip=0pt} 
    \caption{Qualitative comparison of ForgeryTalker and InstructBLIP. For the results, the predicted mask is shown in \textcolor{green}{green} and the ground-truth in \textcolor{red}{red} to highlight localization errors.}
    \label{fig:result}
\end{figure*}

\subsection{Mask Decoder}

We employ SAM's Two-way Transformer \citep{kirillov2023segment} as the mask decoder. The image encoder of InstructBLIP encodes the forgery image. The resulting features from the Q-former are then enhanced through cross-attention with the FPN's regional prompts. These enriched features are subsequently fed into the Two-way Transformer to predict the forgery mask $\hat{M}$. The cross-entropy loss is applied:
\begin{equation}
    \footnotesize
    \mathcal{L}_{m} = -\frac{1}{HW}\sum_{i=1}^{H}\sum_{j=1}^{W}\Big[M_{ij}\log \hat{M}_{ij} + (1-M_{ij})\log(1-\hat{M}_{ij}) \Big],
\end{equation}
where $H, W$ are the height and width of the image. Overall, the full loss in the second stage for {report generation} and forgery localization is formulated as {$\mathcal{L}_{full} = \mathcal{L}_t + \mathcal{L}_m.$}

\section{Experiment}
\label{Experiment}

In this section, we present a series of experiments to evaluate our proposed model, ForgeryTalker, on the MMTT dataset against several baselines.

\subsection{Quantitative Results}

As shown in Table \ref{tab:comparison}, we benchmark our proposed baseline, {ForgeryTalker}, against a comprehensive set of existing models adapted for our task: {SCA} \citep{huang2024segment}, {LISA-7B} \citep{lai2024lisa}, {Osprey} \citep{yuan2024osprey}, {InstructBLIP} \citep{dai2023instructblip}, {FFAA}~\citep{huang2024ffaa}, and {FakeShield}~\citep{xu2024fakeshield}.

\noindent\textbf{Report Generation.} ForgeryTalker obtains the highest CIDEr score ({59.3}) on the standard benchmark, significantly surpassing strong competitors like InstructBLIP (51.7) and LISA-7B (44.1), as well as other baselines including SCA (40.6), Osprey (24.5), FFAA (17.4), and FakeShield (10.0). It also achieves the best performance across all BLEU scores and leads in ROUGE-L ({28.8}). To further verify the generalization capability, we extend the evaluation to the DQ\_F++ dataset~\citep{zhang2024common} in a zero-shot setting. As shown in Table~\ref{tab:dqf_generalization}, ForgeryTalker demonstrates superior performance compared to advanced baselines. Specifically, it achieves a dominant CIDEr score of \textbf{113.3}, significantly outperforming the runner-up InstructBLIP (98.5). Furthermore, ForgeryTalker ranks first in all BLEU metrics, including BLEU-1 (\textbf{48.5}) and BLEU-4 (\textbf{32.4}), while maintaining highly competitive performance on ROUGE-L (\textbf{47.2}). These results validate that our method generalizes effectively to unseen datasets.

\noindent\textbf{Forgery Localization.} ForgeryTalker achieves the highest IoU ({73.67}) and Precision ({91.43}). Its competitive Recall (86.22) is second only to SCA, which achieves the highest Recall of 92.11 but with a notably lower IoU (46.69) and Precision (48.49). Other baselines like InstructBLIP, LISA-7B, and FakeShield report lower IoUs of 64.04, 52.45, and 47.44, respectively. Note that Osprey and FFAA do not provide standalone forgery masks, so their localization metrics are not reported. The qualitative results in Figure~\ref{fig:result} visually corroborate these findings. While InstructBLIP's predicted masks (\textcolor{green}{green}) often over-segment beyond the ground-truth (\textcolor{red}{red}) and its reports are verbose, ForgeryTalker consistently produces more precise masks and concise, relevant reports.

\noindent\textbf{Human Evaluation.} To assess the practical utility of our generated reports to non-experts, we conducted a targeted human evaluation. Five independent evaluators performed a blind test on 100 randomly sampled test images, rating the text on a 1-5 Likert scale for \textit{Faithfulness} (accuracy in identifying artifacts without hallucination) and \textit{Helpfulness} (ability to assist human verification). As shown in Table~\ref{tab:human_eval}, ForgeryTalker significantly outperforms baselines, demonstrating that explicitly grounding text in localized visual evidence substantially calibrates human trust.

\begin{table}[t!]
\centering
\resizebox{\columnwidth}{!}{%
\begin{tabular}{lcc}
\toprule[1.5pt]
\textbf{Method} & \textbf{Faithfulness $\uparrow$} & \textbf{Helpfulness $\uparrow$} \\
\hline
SCA~\citep{huang2024segment} & 2.3 & 3.1 \\
LISA-7B~\citep{lai2024lisa} & 2.7 & 3.4 \\
InstructBLIP~\citep{dai2023instructblip} & 3.6 & 3.8 \\
\rowcolor{gray!15} \textbf{ForgeryTalker} & \textbf{4.3} & \textbf{4.4} \\
\bottomrule[1.5pt]
\end{tabular}%
}
\vspace{-0.2cm}
\caption{Human Evaluation results (Average Scores out of 5.0) assessing the faithfulness and helpfulness of generated reports.}
\label{tab:human_eval}
\vspace{-0.2cm}
\end{table}

\begin{table*}[t!]
\centering
\footnotesize
\begin{tabular}{lcccccc}
\toprule[1.5pt] 
\textbf{Model} & \textbf{Bleu\_1} & \textbf{Bleu\_2} & \textbf{Bleu\_3} & \textbf{Bleu\_4} & \textbf{ROUGE\_L} & \textbf{CIDEr} \\
\hline
SCA~\citep{huang2024segment}  & 47.6 & 39.9 & 35.2 & 30.1 & 40.4 & 71.0 \\
LISA-7B~\citep{lai2024lisa} & 46.5 & 38.4 & 33.1 & 31.2 & 45.6 & 74.3 \\
InstructBLIP~\citep{dai2023instructblip} & 43.5 & 38.0 & 34.2 & 31.0 & 47.9 & 98.5 \\
ForgeryTalker & \textbf{48.5} & \textbf{41.4} & \textbf{36.5} & \textbf{32.4} & \textbf{47.2} & \textbf{113.3} \\
\bottomrule[1.5pt]
\end{tabular}
\vspace{-0.2cm}
\caption{Report generation comparison on the DQ\_F++ dataset~\citep{zhang2024common}. The best score for each metric is shown in \textbf{bold}.}
\label{tab:dqf_generalization}
\vspace{-0.2cm}
\end{table*}

\begin{table*}[t!]
\centering
\vspace{-.1in}
\resizebox{.95\textwidth}{!}{%
\begin{tabular}{lcccccccccc}
\toprule[1.5pt]
                \multirow{2}{*}{\textbf{Method}} & \multicolumn{6}{c}{\textbf{Report Generation}} & \multicolumn{3}{c}{\textbf{Forgery Localization}} \\
                \cmidrule(r){2-7} \cmidrule(r){8-10}
                & {CIDEr} & {Bleu\_1} & {Bleu\_2} & {Bleu\_3} & {Bleu\_4}  & {ROUGE\_L} & {IoU}  & Precision & Recall \\ 
\hline
ForgeryTalker $w/$ FPN-GT & \textbf{95.1} & \textbf{41.5} & \textbf{27.6} & \textbf{20.3} & \textbf{16.0} & \textbf{37.0} & 66.90 & 88.74 & 79.83\\ 
ForgeryTalker $w/o$ FPN & 51.7 & 31.8 & 20.3 & 14.6 & 11.4 & 27.7 & 64.04 & 87.88 & 78.53\\ \hline
ForgeryTalker & 59.3 & 35.0 & 22.1 & 16.0 & 12.5 & 28.8 & \textbf{73.67} & \textbf{91.43} & \textbf{86.22}\\ 
\bottomrule[1.5pt]
\end{tabular}}
\vspace{-.1in}
\caption{Ablation study on the impact of different variants. $w/$ and $w/o$ mean equipping or not equipping the following modules.}
\label{tab:AblationStudy}
\end{table*}

\begin{table*}[t!]
    \captionsetup{skip=3pt} 
    \vspace{-.1in}
    \begin{center}
        \resizebox{0.9\textwidth}{!}{ 
            \begin{tabular}{lcccccccccc}
            \toprule[1.5pt]
                \multirow{2}{*}{\textbf{Method}} 
                & \multicolumn{6}{c}{\textbf{Report Generation}} 
                & \multicolumn{3}{c}{\textbf{Forgery Localization}} \\
                \cmidrule(r){2-7} \cmidrule(r){8-10}
                & {CIDEr} & {BLEU-1} & {BLEU-2} & {BLEU-3} & {BLEU-4} & {ROUGE-L}
                & {IoU}  & {Precision} & {Recall}\\
            \midrule
            $w/o$ Pretraining Stage 
                & 54.4 & 33.8 & 21.6 & 15.5 & 12.1 & 28.3 
                & 65.87 & 89.00 & 78.87 \\ 
            \hline
            \textbf{$w/$ Pretraining Stage} & \textbf{59.3} & \textbf{35.0} & \textbf{22.1} & \textbf{16.0} & \textbf{12.5} & \textbf{28.8} & 73.67 & \textbf{91.43} & \textbf{86.22}\\
            \bottomrule[1.5pt]
            \end{tabular}
        }
    \end{center}
    \vspace{-0.2cm}
    \caption{Impact of the Forgery-aware Pretraining Stage.}
    \label{tab:pretrainloss}
    \vspace{-0.2cm}
\end{table*}

\subsection{Ablation Study}

\noindent\textbf{Effect of the Forgery Prompter Network (FPN).} The FPN is shown to be a critical component. Table~\ref{tab:AblationStudy} reveals a significant performance drop in report generation (CIDEr drops to 51.7) when the FPN is removed. Conversely, an oracle FPN using ground-truth prompts (w/ FPN-GT) establishes a high upper bound at 95.1 CIDEr. This large performance gap underscores that the quality of region prompts is a key factor for this task and motivates future work on improving the FPN module.

\noindent\textbf{Effectiveness of Pretraining Stage.} Table~\ref{tab:pretrainloss} empirically validates the critical importance of our forgery-aware pretraining strategy. Compared to the baseline model trained purely from scratch, integrating the pretraining stage yields substantial improvements across all evaluated metrics. Specifically, it significantly boosts the localization performance (\eg, IoU +7.8\%) and enhances the semantic quality of generated reports (\eg, CIDEr +4.9). This confirms that optimizing the four pretraining objectives equips the model with robust multimodal representations, which are pivotal for achieving high performance on the downstream joint task.

\begin{table*}[t!]
\centering
\footnotesize
\resizebox{0.9\textwidth}{!}{
\begin{tabular}{lccccccc}
\toprule[1.5pt]
\textbf{Degradation Condition} & \textbf{CIDEr} & \textbf{Bleu\_1} & \textbf{Bleu\_4} & \textbf{ROUGE\_L} & \textbf{IoU} & \textbf{Precision} & \textbf{Recall} \\
\hline
Clean (Original) & \textbf{59.3} & \textbf{35.0} & \textbf{12.5} & \textbf{28.8} & 73.67 & \textbf{91.43} & \textbf{86.22} \\
\midrule
Moderate ($0.75\times$ Down-sampling) & 59.1 & 34.8 & 12.4 & 28.8 & 74.31 & 84.12 & 84.68 \\
Moderate (Kernel 5 Blur) & 44.8 & 31.9 & 8.8 & 26.2 & 65.89 & 84.50 & 74.09 \\
\midrule
Severe ($0.5\times$ Down-sampling) & 59.7 & 35.1 & 12.5 & 28.8 & \textbf{74.87} & 84.22 & 85.14 \\
Severe (Kernel 11 Blur) & 39.1 & 31.3 & 7.8 & 25.4 & 59.51 & 80.82 & 69.44 \\
\bottomrule[1.5pt]
\end{tabular}}
\vspace{-0.2cm}
\caption{Robustness evaluation across various degradation levels simulating in-the-wild scenarios.}
\label{tab:robustness}
\vspace{-0.4cm}
\end{table*}

\noindent\textbf{Impact of Freezing the LLM.} Keeping the LLM frozen is a foundational practice in our design to preserve its pre-trained linguistic reasoning and prevent catastrophic forgetting. Empirical results validate this strategy: fine-tuning the LLM with LoRA degrades the text generation performance (dropping from 54.4 to 49.6 CIDEr) compared to the frozen setting. This confirms that preserving the LLM's pre-trained integrity while tuning alignment modules (e.g., Q-Former and FPN) is optimal for forgery attribution.

\noindent\textbf{Robustness to Real-World Degradations.} 
Images in the wild frequently suffer from social media compression and blurring. To evaluate robustness, we tested ForgeryTalker under varying degradation levels (Table~\ref{tab:robustness}). Remarkably, under severe $0.5\times$ down-sampling, the CIDEr score remains stable (59.7), and the mask IoU slightly increases to 74.87\%, proving the model leverages scale-invariant structural cues rather than fragile pixel-level noise. While severe Gaussian blur (Kernel 11) physically erases high-frequency forensic details and impacts metrics (reducing CIDEr to 39.1), the model still maintains functional semantic reasoning based on global patterns.

\section{Conclusion}

We address the limitations of traditional forgery localization methods, which typically lack explanatory power. We introduce the novel task of \textbf{Forgery Attribution Report Generation} to produce both precise localization masks and rich textual explanations. To catalyze this research, we release the \textbf{MMTT} dataset, the first large-scale benchmark featuring high-precision, process-derived masks and meticulously crafted annotations. Furthermore, we propose \textbf{ForgeryTalker}, a unified baseline that integrates localization and report generation into an end-to-end framework. Our experiments validate the effectiveness of ForgeryTalker and establish a solid benchmark on the MMTT dataset. These contributions pave the way for future advancements in explainable and trustworthy facial forgery analysis.



\section*{Acknowledgements}

This work is supported by the National Natural Science Foundation of China (Grant No. 62302140), the National Key Research and Development Program of China (Grant No. 2023YFC3321600), the Guangdong Basic and Applied Basic Research Foundation (Grant No. 2025A1515012281), the Jiangsu Provincial Science and Technology Program (Grant No. SBZ20250900116), and the Macao Science and Technology Development Fund (Grant No. FDCT/0043/2025/RIA1). Finally, we thank the anonymous reviewers and area chairs for their constructive feedback, which helped improve this paper.

\section*{Limitations}

Despite the promising performance of ForgeryTalker and the MMTT dataset, we acknowledge several limitations. 
First, our framework provides post-hoc explanations based on the premise of an upstream binary decision. As we focus on "where" and "why" for flagged images, the model does not collect interpretable local evidence to \textit{drive} the initial detection. Transitioning from post-hoc attribution to a fully integrated, evidence-based detection pipeline remains an important open research direction.
Second, the quality of the generated reports is highly sensitive to the accuracy of initial region proposals. As shown in our ablation studies, while the Forgery Prompter Network (FPN) establishes a robust baseline, there remains a performance gap compared to guidance using ground-truth masks. Bridging this bottleneck is crucial for more reliable forensic reporting.
Finally, while ForgeryTalker is efficient during inference, the two-stage training pipeline (forgery-aware pre-training and task-specific fine-tuning) incurs higher computational costs compared to training simpler models from scratch. Additionally, although MMTT covers major manipulation paradigms, potential synthetic bias remains a challenge when encountering entirely unseen, adversarial real-world forgeries.

\section*{Ethical Considerations}
The proposed \textbf{MMTT} dataset and the associated \textbf{ForgeryTalker} framework are developed exclusively to support academic research on deepfake detection and interpretable forensics. We acknowledge the potential dual-use risks inherent in constructing high-fidelity forged facial imagery, which could be misused to analyze or circumvent detection systems. To mitigate such risks, we adopt a strict harm-minimization and controlled-release policy. Specifically, we do \textbf{not} disclose the detailed generation pipelines or adversarial editing tools to prevent direct exploitation by malicious actors. Access to the dataset is restricted to vetted academic researchers and institutions under a signed Data Usage Agreement (DUA), which explicitly prohibits malicious content generation and identity re-identification. Moreover, all source images are obtained from publicly available academic datasets and are carefully screened to exclude minors. We reserve the right to revoke access upon any evidence of misuse.

\bibliography{custom}

\begin{thebibliography}{64}
\providecommand{\natexlab}[1]{#1}

\bibitem[{Abou~Akar et~al.(2024)Abou~Akar, Abdel~Massih, Yaghi, Khalil, Kamradt, and Makhoul}]{e4s}
Chafic Abou~Akar, Rachelle Abdel~Massih, Anthony Yaghi, Joe Khalil, Marc Kamradt, and Abdallah Makhoul. 2024.
\newblock Generative adversarial network applications in industry 4.0: A review.
\newblock \emph{International Journal of Computer Vision}, 132(6):2195--2254.

\bibitem[{Alexey(2020)}]{alexey2020image}
Dosovitskiy Alexey. 2020.
\newblock An image is worth 16x16 words: Transformers for image recognition at scale.
\newblock \emph{arXiv preprint arXiv: 2010.11929}.

\bibitem[{Anan et~al.(2025)Anan, Bhattacharjee, Intesher, Islam, Assaeem~Fuad, Saha, and Imtiaz}]{anan2025hybrid}
Kafi Anan, Anindya Bhattacharjee, Ashir Intesher, Kaidul Islam, Abrar Assaeem~Fuad, Utsab Saha, and Hafiz Imtiaz. 2025.
\newblock Hybrid deepfake image detection: A comprehensive dataset-driven approach integrating convolutional and attention mechanisms with frequency domain features.
\newblock \emph{arXiv e-prints}, pages arXiv--2502.

\bibitem[{Bai et~al.(2025)Bai, Chen, Liu, Wang, Ge, Song, Dang, Wang, Wang, Tang, Zhong, Zhu, Yang, Li, Wan, Wang, Ding, Fu, Xu, Ye, Zhang, Xie, Cheng, Zhang, Yang, Xu, and Lin}]{bai2025qwen25vltechnicalreport}
Shuai Bai, Keqin Chen, Xuejing Liu, Jialin Wang, Wenbin Ge, Sibo Song, Kai Dang, Peng Wang, Shijie Wang, Jun Tang, Humen Zhong, Yuanzhi Zhu, Mingkun Yang, Zhaohai Li, Jianqiang Wan, Pengfei Wang, Wei Ding, Zheren Fu, Yiheng Xu, and 8 others. 2025.
\newblock \href {https://arxiv.org/abs/2502.13923} {Qwen2.5-vl technical report}.
\newblock \emph{Preprint}, arXiv:2502.13923.

\bibitem[{Chen et~al.(2024)Chen, Sun, Zhou, Lin, Sun, Cao, and Ji}]{chen2024diffusionface}
Zhongxi Chen, Ke~Sun, Ziyin Zhou, Xianming Lin, Xiaoshuai Sun, Liujuan Cao, and Rongrong Ji. 2024.
\newblock Diffusionface: Towards a comprehensive dataset for diffusion-based face forgery analysis.
\newblock \emph{arXiv preprint arXiv:2403.18471}.

\bibitem[{Chung et~al.(2022)Chung, Hou, Longpre, Zoph, Tay, Fedus, Li, Wang, Dehghani, Brahma, Webson, Gu, Dai, Suzgun, Chen, Chowdhery, Castro-Ros, Pellat, Robinson, Valter, Narang, Mishra, Yu, Zhao, Huang, Dai, Yu, Petrov, Chi, Dean, Devlin, Roberts, Zhou, Le, and Wei}]{chung2022scaling}
Hyung~Won Chung, Le~Hou, Shayne Longpre, Barret Zoph, Yi~Tay, William Fedus, Yunxuan Li, Xuezhi Wang, Mostafa Dehghani, Siddhartha Brahma, Albert Webson, Shixiang~Shane Gu, Zhuyun Dai, Mirac Suzgun, Xinyun Chen, Aakanksha Chowdhery, Alex Castro-Ros, Marie Pellat, Kevin Robinson, and 16 others. 2022.
\newblock \href {https://arxiv.org/abs/2210.11416} {Scaling instruction-finetuned language models}.
\newblock \emph{Preprint}, arXiv:2210.11416.

\bibitem[{Dai et~al.(2023)Dai, Li, Li, Tiong, Zhao, Wang, Li, Fung, and Hoi}]{dai2023instructblip}
Wenliang Dai, Junnan Li, Dongxu Li, Anthony Meng~Huat Tiong, Junqi Zhao, Weisheng Wang, Boyang Li, Pascale Fung, and Steven Hoi. 2023.
\newblock \href {https://arxiv.org/abs/2305.06500} {Instructblip: Towards general-purpose vision-language models with instruction tuning}.
\newblock \emph{Preprint}, arXiv:2305.06500.

\bibitem[{Dhariwal and Nichol(2021)}]{dhariwal2021diffusion}
Prafulla Dhariwal and Alexander Nichol. 2021.
\newblock Diffusion models beat gans on image synthesis.
\newblock \emph{Advances in neural information processing systems}, 34:8780--8794.

\bibitem[{Dolhansky et~al.(2020)Dolhansky, Bitton, Pflaum, Lu, Howes, Wang, and Ferrer}]{dolhansky2020deepfake}
Brian Dolhansky, Joanna Bitton, Ben Pflaum, Jikuo Lu, Russ Howes, Menglin Wang, and Cristian~Canton Ferrer. 2020.
\newblock The deepfake detection challenge (dfdc) dataset.
\newblock \emph{arXiv preprint arXiv:2006.07397}.

\bibitem[{Guo et~al.(2025)Guo, Wu, Zhu, Leng, Shi, Chen, Fan, Wang, Jiang, Wang, Chen, Huang, Lei, Yuan, Luo, Liu, Ye, Qian, Yan, Zhao, Peng, Li, Yuan, Wu, Cheng, Liu, Wang, Zeng, Liu, Qin, Ding, Xiao, Zhang, Zhang, Xiong, Peng, Chen, Li, Hu, Lin, Hu, Zhang, Wu, Li, Liu, Ling, Qin, Wang, He, Zhang, Yi, Liao, Huang, Zhang, Deng, Deng, Lin, Yuan, Li, Gou, Lou, Wei, Liu, Li, Zhu, Zhong, Li, Zhang, Wu, Li, Xiao, Lin, Yang, Wang, Ji, Hao, Shen, Li, Li, Wu, Zhu, Jiao, Feng, Chen, Duan, Liu, Zeng, Tang, Sun, Chen, Long, Feng, Zhan, Fang, Lu, Hua, Liu, Shen, Zhang, Shen, Wang, Pan, Zhang, Li, Li, Li, Shi, Han, Xiang, Chen, Chen, Li, Yan, Chi, Liu, Du, Wang, Pan, Chen, Chen, Wu, Yuan, Shuai, Tao, Zheng, Zhang, Zhang, Wang, Yang, Zhao, Xu, Liang, Yan, Zhong, Cao, Wu, Liu, Chang, Cai, Ao, Yang, Zhang, Zhong, Jia, Weng, Yu, Huang, Zhu, Yang, Wang, Long, Yin, Li, Zhu, Jia, Zhang, Liu, Zhang, Yang, Luo, Chen, Zhong, Xiao, Li, Wu, Wen, Du, Zhang, Ye, Wu, Liu, Yue, Zhou, Yuan, Xu, Yang, Zhang, Fang, Li, Ren, Xiong, Hong,
  Wang, Sun, Wang, Cai, Zha, An, Zhao, Xu, Chen, Wu, Zheng, Wang, Huang, Zhu, and Song}]{guo2025seed15vltechnicalreport}
Dong Guo, Faming Wu, Feida Zhu, Fuxing Leng, Guang Shi, Haobin Chen, Haoqi Fan, Jian Wang, Jianyu Jiang, Jiawei Wang, Jingji Chen, Jingjia Huang, Kang Lei, Liping Yuan, Lishu Luo, Pengfei Liu, Qinghao Ye, Rui Qian, Shen Yan, and 178 others. 2025.
\newblock \href {https://arxiv.org/abs/2505.07062} {Seed1.5-vl technical report}.
\newblock \emph{Preprint}, arXiv:2505.07062.

\bibitem[{He et~al.(2021)He, Gan, Chen, Zhou, Yin, Song, Sheng, Shao, and Liu}]{he2021forgerynet}
Yinan He, Bei Gan, Siyu Chen, Yichun Zhou, Guojun Yin, Luchuan Song, Lu~Sheng, Jing Shao, and Ziwei Liu. 2021.
\newblock Forgerynet: A versatile benchmark for comprehensive forgery analysis.
\newblock In \emph{Proceedings of the IEEE/CVF conference on computer vision and pattern recognition}, pages 4360--4369.

\bibitem[{Ho et~al.(2020)Ho, Jain, and Abbeel}]{ho2020denoising}
Jonathan Ho, Ajay Jain, and Pieter Abbeel. 2020.
\newblock Denoising diffusion probabilistic models.
\newblock \emph{Advances in neural information processing systems}, 33:6840--6851.

\bibitem[{Huang et~al.(2024{\natexlab{a}})Huang, Wang, Tang, Zhang, Hu, Lu, Wang, and Liu}]{huang2024segment}
Xiaoke Huang, Jianfeng Wang, Yansong Tang, Zheng Zhang, Han Hu, Jiwen Lu, Lijuan Wang, and Zicheng Liu. 2024{\natexlab{a}}.
\newblock Segment and caption anything.
\newblock In \emph{Proceedings of the IEEE/CVF Conference on Computer Vision and Pattern Recognition}, pages 13405--13417.

\bibitem[{Huang et~al.(2024{\natexlab{b}})Huang, Xia, Lin, Mou, Yang, and Jia}]{huang2024ffaa}
Zhengchao Huang, Bin Xia, Zicheng Lin, Zhun Mou, Wenming Yang, and Jiaya Jia. 2024{\natexlab{b}}.
\newblock Ffaa: Multimodal large language model based explainable open-world face forgery analysis assistant.
\newblock \emph{arXiv preprint arXiv:2408.10072}.

\bibitem[{Jiang et~al.(2020)Jiang, Li, Wu, Qian, and Loy}]{jiang2020deeperforensics}
Liming Jiang, Ren Li, Wayne Wu, Chen Qian, and Chen~Change Loy. 2020.
\newblock Deeperforensics-1.0: A large-scale dataset for real-world face forgery detection.
\newblock In \emph{Proceedings of the IEEE/CVF conference on computer vision and pattern recognition}, pages 2889--2898.

\bibitem[{Kaddar et~al.(2021)Kaddar, Fezza, Hamidouche, Akhtar, and Hadid}]{kaddar2021hcit}
Bachir Kaddar, Sid~Ahmed Fezza, Wassim Hamidouche, Zahid Akhtar, and Abdenour Hadid. 2021.
\newblock Hcit: Deepfake video detection using a hybrid model of cnn features and vision transformer.
\newblock In \emph{2021 International Conference on Visual Communications and Image Processing (VCIP)}, pages 1--5. IEEE.

\bibitem[{Kang et~al.(2025)Kang, Wen, Wen, Ye, Li, Feng, Zhou, Wang, Lin, Zhang, and He}]{kang2025legionlearninggroundexplain}
Hengrui Kang, Siwei Wen, Zichen Wen, Junyan Ye, Weijia Li, Peilin Feng, Baichuan Zhou, Bin Wang, Dahua Lin, Linfeng Zhang, and Conghui He. 2025.
\newblock \href {https://arxiv.org/abs/2503.15264} {Legion: Learning to ground and explain for synthetic image detection}.
\newblock \emph{Preprint}, arXiv:2503.15264.

\bibitem[{Karras et~al.(2019)Karras, Laine, and Aila}]{karras2019style}
Tero Karras, Samuli Laine, and Timo Aila. 2019.
\newblock A style-based generator architecture for generative adversarial networks.
\newblock In \emph{Proceedings of the IEEE/CVF conference on computer vision and pattern recognition}, pages 4401--4410.

\bibitem[{Khedkar et~al.(2022)Khedkar, Peshkar, Nagdive, Gaikwad, and Baudha}]{khedkar2022exploiting}
Atharva Khedkar, Atharva Peshkar, Ashlesha Nagdive, Mahendra Gaikwad, and Sudeep Baudha. 2022.
\newblock Exploiting spatiotemporal inconsistencies to detect deepfake videos in the wild.
\newblock In \emph{2022 10th International Conference on Emerging Trends in Engineering and Technology-Signal and Information Processing (ICETET-SIP-22)}, pages 1--6. IEEE.

\bibitem[{Kirillov et~al.(2023)Kirillov, Mintun, Ravi, Mao, Rolland, Gustafson, Xiao, Whitehead, Berg, Lo et~al.}]{kirillov2023segment}
Alexander Kirillov, Eric Mintun, Nikhila Ravi, Hanzi Mao, Chloe Rolland, Laura Gustafson, Tete Xiao, Spencer Whitehead, Alexander~C Berg, Wan-Yen Lo, and 1 others. 2023.
\newblock Segment anything.
\newblock In \emph{Proceedings of the IEEE/CVF International Conference on Computer Vision}, pages 4015--4026.

\bibitem[{Lai et~al.(2024)Lai, Tian, Chen, Li, Yuan, Liu, and Jia}]{lai2024lisa}
Xin Lai, Zhuotao Tian, Yukang Chen, Yanwei Li, Yuhui Yuan, Shu Liu, and Jiaya Jia. 2024.
\newblock Lisa: Reasoning segmentation via large language model.
\newblock In \emph{Proceedings of the IEEE/CVF Conference on Computer Vision and Pattern Recognition}, pages 9579--9589.

\bibitem[{Lalitha and Sooda(2022)}]{lalitha2022deepfake}
S~Lalitha and Kavitha Sooda. 2022.
\newblock Deepfake detection through key video frame extraction using gan.
\newblock In \emph{2022 International Conference on Automation, Computing and Renewable Systems (ICACRS)}, pages 859--863. IEEE.

\bibitem[{Le et~al.(2021)Le, Nguyen, Yamagishi, and Echizen}]{le2021openforensics}
Trung-Nghia Le, Huy~H Nguyen, Junichi Yamagishi, and Isao Echizen. 2021.
\newblock Openforensics: Large-scale challenging dataset for multi-face forgery detection and segmentation in-the-wild.
\newblock In \emph{Proceedings of the IEEE/CVF international conference on computer vision}, pages 10117--10127.

\bibitem[{Li et~al.(2019)Li, Bao, Yang, Chen, and Wen}]{li2019faceshifter}
Lingzhi Li, Jianmin Bao, Hao Yang, Dong Chen, and Fang Wen. 2019.
\newblock Faceshifter: Towards high fidelity and occlusion aware face swapping.
\newblock \emph{arXiv preprint arXiv:1912.13457}.

\bibitem[{Li et~al.(2022)Li, Lin, Zhou, Qi, Wang, and Jia}]{mat}
Wenbo Li, Zhe Lin, Kun Zhou, Lu~Qi, Yi~Wang, and Jiaya Jia. 2022.
\newblock Mat: Mask-aware transformer for large hole image inpainting.
\newblock In \emph{Proceedings of the IEEE/CVF conference on computer vision and pattern recognition}, pages 10758--10768.

\bibitem[{Li et~al.(2020)Li, Yang, Sun, Qi, and Lyu}]{li2020celeb}
Yuezun Li, Xin Yang, Pu~Sun, Honggang Qi, and Siwei Lyu. 2020.
\newblock Celeb-df: A large-scale challenging dataset for deepfake forensics.
\newblock In \emph{Proceedings of the IEEE/CVF conference on computer vision and pattern recognition}, pages 3207--3216.

\bibitem[{Liu et~al.(2022{\natexlab{a}})Liu, Yang, Zhang, and Liu}]{liu2022maskgan}
Dazhuang Liu, Zhen Yang, Ru~Zhang, and Jianyi Liu. 2022{\natexlab{a}}.
\newblock Maskgan: A facial fusion algorithm for deepfake image detection.
\newblock In \emph{2022 International Conference on Computers and Artificial Intelligence Technologies (CAIT)}, pages 71--78. IEEE.

\bibitem[{Liu et~al.(2024)Liu, Li, Li, Li, Zhang, Shen, and Lee}]{liu2024llavanext}
Haotian Liu, Chunyuan Li, Yuheng Li, Bo~Li, Yuanhan Zhang, Sheng Shen, and Yong~Jae Lee. 2024.
\newblock \href {https://llava-vl.github.io/blog/2024-01-30-llava-next/} {Llava-next: Improved reasoning, ocr, and world knowledge}.

\bibitem[{Liu et~al.(2023{\natexlab{a}})Liu, Wang, and Li}]{liu2023interpretable}
Hui Liu, Wenya Wang, and Haoliang Li. 2023{\natexlab{a}}.
\newblock Interpretable multimodal misinformation detection with logic reasoning.
\newblock In \emph{Findings of the Association for Computational Linguistics: ACL 2023}, pages 9781--9796.

\bibitem[{Liu et~al.(2023{\natexlab{b}})Liu, Perov, Gao, Chervoniy, Zhou, and Zhang}]{liu2023deepfacelab}
Kunlin Liu, Ivan Perov, Daiheng Gao, Nikolay Chervoniy, Wenbo Zhou, and Weiming Zhang. 2023{\natexlab{b}}.
\newblock Deepfacelab: Integrated, flexible and extensible face-swapping framework.
\newblock \emph{Pattern Recognition}, 141:109628.

\bibitem[{Liu et~al.(2025{\natexlab{a}})Liu, Wang, Zhu, and Zheng}]{liu2025every}
Lingyu Liu, Yaxiong Wang, Li~Zhu, and Zhedong Zheng. 2025{\natexlab{a}}.
\newblock Every painting awakened: A training-free framework for painting-to-animation generation.
\newblock \emph{arXiv preprint arXiv:2503.23736}.

\bibitem[{Liu et~al.(2018)Liu, Lehman, Molino, Petroski~Such, Frank, Sergeev, and Yosinski}]{liu2018intriguing}
Rosanne Liu, Joel Lehman, Piero Molino, Felipe Petroski~Such, Eric Frank, Alex Sergeev, and Jason Yosinski. 2018.
\newblock An intriguing failing of convolutional neural networks and the coordconv solution.
\newblock \emph{Advances in neural information processing systems}, 31.

\bibitem[{Liu et~al.(2022{\natexlab{b}})Liu, Liu, Chen, and Liu}]{liu2022pscc}
Xiaohong Liu, Yaojie Liu, Jun Chen, and Xiaoming Liu. 2022{\natexlab{b}}.
\newblock Pscc-net: Progressive spatio-channel correlation network for image manipulation detection and localization.
\newblock \emph{IEEE Transactions on Circuits and Systems for Video Technology}, 32(11):7505--7517.

\bibitem[{Liu et~al.(2025{\natexlab{b}})Liu, Xiao, , Liu, Bengio, and Zhang}]{liu2025nablagfn}
Zhen Liu, Tim~Z Xiao, , Weiyang Liu, Yoshua Bengio, and Dinghuai Zhang. 2025{\natexlab{b}}.
\newblock Efficient diversity-preserving diffusion alignment via gradient-informed gflownets.
\newblock In \emph{ICLR}.

\bibitem[{Livernoche et~al.(2025)Livernoche, Arodi, Musulan, Yang, Salvail, Caron, Godbout, and Rabbany}]{livernoche2025openfake}
Victor Livernoche, Akshatha Arodi, Andreea Musulan, Zachary Yang, Adam Salvail, Ga{\'e}tan~Marceau Caron, Jean-Fran{\c{c}}ois Godbout, and Reihaneh Rabbany. 2025.
\newblock Openfake: An open dataset and platform toward real-world deepfake detection.
\newblock \emph{arXiv preprint arXiv:2509.09495}.

\bibitem[{Ma et~al.(2023)Ma, Du, Jiang, Hammadi, and Zhou}]{ma2023imlvit}
Xiaochen Ma, Bo~Du, Zhuohang Jiang, Ahmed Y.~Al Hammadi, and Jizhe Zhou. 2023.
\newblock \href {https://arxiv.org/abs/2307.14863} {Iml-vit: Benchmarking image manipulation localization by vision transformer}.
\newblock \emph{Preprint}, arXiv:2307.14863.

\bibitem[{Ma et~al.(2024)Ma, Luo, Guo, Zeng, Hao, and Zhao}]{ma2024event}
Zihan Ma, Minnan Luo, Hao Guo, Zhi Zeng, Yiran Hao, and Xiang Zhao. 2024.
\newblock Event-radar: Event-driven multi-view learning for multimodal fake news detection.
\newblock In \emph{Proceedings of the 62nd Annual Meeting of the Association for Computational Linguistics (Volume 1: Long Papers)}, pages 5809--5821.

\bibitem[{Neves et~al.(2020)Neves, Tolosana, Vera-Rodriguez, Lopes, Proen{\c{c}}a, and Fierrez}]{neves2020ganprintr}
Joao~C Neves, Ruben Tolosana, Ruben Vera-Rodriguez, Vasco Lopes, Hugo Proen{\c{c}}a, and Julian Fierrez. 2020.
\newblock Ganprintr: Improved fakes and evaluation of the state of the art in face manipulation detection.
\newblock \emph{IEEE Journal of Selected Topics in Signal Processing}, 14(5):1038--1048.

\bibitem[{Paszke et~al.(2019)Paszke, Gross, Massa, Lerer, Bradbury, Chanan, Killeen, Lin, Gimelshein, Antiga, Desmaison, K{\"{o}}pf, Yang, DeVito, Raison, Tejani, Chilamkurthy, Steiner, Fang, Bai, and Chintala}]{pytorch}
Adam Paszke, Sam Gross, Francisco Massa, Adam Lerer, James Bradbury, Gregory Chanan, Trevor Killeen, Zeming Lin, Natalia Gimelshein, Luca Antiga, Alban Desmaison, Andreas K{\"{o}}pf, Edward~Z. Yang, Zachary DeVito, Martin Raison, Alykhan Tejani, Sasank Chilamkurthy, Benoit Steiner, Lu~Fang, and 2 others. 2019.
\newblock \href {https://proceedings.neurips.cc/paper/2019/hash/bdbca288fee7f92f2bfa9f7012727740-Abstract.html} {Pytorch: An imperative style, high-performance deep learning library}.
\newblock In \emph{Advances in Neural Information Processing Systems 32: Annual Conference on Neural Information Processing Systems 2019, NeurIPS 2019, December 8-14, 2019, Vancouver, BC, Canada}, pages 8024--8035.

\bibitem[{Patashnik et~al.(2021)Patashnik, Wu, Shechtman, Cohen-Or, and Lischinski}]{Patashnik_2021_ICCV}
Or~Patashnik, Zongze Wu, Eli Shechtman, Daniel Cohen-Or, and Dani Lischinski. 2021.
\newblock Styleclip: Text-driven manipulation of stylegan imagery.
\newblock In \emph{Proceedings of the IEEE/CVF International Conference on Computer Vision (ICCV)}, pages 2085--2094.

\bibitem[{Podell et~al.(2023)Podell, English, Lacey, Blattmann, Dockhorn, M{\"u}ller, Penna, and Rombach}]{sdxl}
Dustin Podell, Zion English, Kyle Lacey, Andreas Blattmann, Tim Dockhorn, Jonas M{\"u}ller, Joe Penna, and Robin Rombach. 2023.
\newblock Sdxl: Improving latent diffusion models for high-resolution image synthesis.
\newblock \emph{arXiv preprint arXiv:2307.01952}.

\bibitem[{Ramachandran et~al.(2021)Ramachandran, Nadimpalli, and Rattani}]{ramachandran2021experimental}
Sreeraj Ramachandran, Aakash~Varma Nadimpalli, and Ajita Rattani. 2021.
\newblock An experimental evaluation on deepfake detection using deep face recognition.
\newblock In \emph{2021 International Carnahan Conference on Security Technology (ICCST)}, pages 1--6. IEEE.

\bibitem[{Rana et~al.(2022)Rana, Nobi, Murali, and Sung}]{rana2022deepfake}
Md~Shohel Rana, Mohammad~Nur Nobi, Beddhu Murali, and Andrew~H Sung. 2022.
\newblock Deepfake detection: A systematic literature review.
\newblock \emph{IEEE access}, 10:25494--25513.

\bibitem[{Richards et~al.(2023)Richards, Varshini, Diviya, Prakash, Kasthuri, and Sasithradevi}]{richards2023deep}
Mj~Alben Richards, E~Kaaviya Varshini, N~Diviya, P~Prakash, P~Kasthuri, and A~Sasithradevi. 2023.
\newblock Deep fake face detection using convolutional neural networks.
\newblock In \emph{2023 12th International Conference on Advanced Computing (ICoAC)}, pages 1--5. IEEE.

\bibitem[{Ross and Doll{\'a}r(2017)}]{ross2017focal}
T-YLPG Ross and GKHP Doll{\'a}r. 2017.
\newblock Focal loss for dense object detection.
\newblock In \emph{proceedings of the IEEE conference on computer vision and pattern recognition}, pages 2980--2988.

\bibitem[{Rossler et~al.(2019)Rossler, Cozzolino, Verdoliva, Riess, Thies, and Nie{\ss}ner}]{rossler2019faceforensics++}
Andreas Rossler, Davide Cozzolino, Luisa Verdoliva, Christian Riess, Justus Thies, and Matthias Nie{\ss}ner. 2019.
\newblock Faceforensics++: Learning to detect manipulated facial images.
\newblock In \emph{Proceedings of the IEEE/CVF international conference on computer vision}, pages 1--11.

\bibitem[{Sabir et~al.(2019)Sabir, Cheng, Jaiswal, AbdAlmageed, Masi, and Natarajan}]{sabir2019recurrent}
Ekraam Sabir, Jiaxin Cheng, Ayush Jaiswal, Wael AbdAlmageed, Iacopo Masi, and Prem Natarajan. 2019.
\newblock Recurrent convolutional strategies for face manipulation detection in videos.
\newblock \emph{Interfaces (GUI)}, 3(1):80--87.

\bibitem[{Song et~al.(2020)Song, Meng, and Ermon}]{song2020denoising}
Jiaming Song, Chenlin Meng, and Stefano Ermon. 2020.
\newblock Denoising diffusion implicit models.
\newblock \emph{arXiv preprint arXiv:2010.02502}.

\bibitem[{Sun et~al.(2023)Sun, Zhang, Echizen, Nguyen, Qiu, and Sun}]{sun2023face}
YuYang Sun, ZhiYong Zhang, Isao Echizen, Huy~H Nguyen, ChangZhen Qiu, and Lu~Sun. 2023.
\newblock Face forgery detection based on facial region displacement trajectory series.
\newblock In \emph{Proceedings of the IEEE/CVF Winter Conference on Applications of Computer Vision}, pages 633--642.

\bibitem[{Verdoliva(2020)}]{verdoliva2020media}
Luisa Verdoliva. 2020.
\newblock Media forensics and deepfakes: an overview.
\newblock \emph{IEEE journal of selected topics in signal processing}, 14(5):910--932.

\bibitem[{Wang et~al.(2022)Wang, Zhang, Fan, Wang, and Chen}]{wang2021HFGI}
Tengfei Wang, Yong Zhang, Yanbo Fan, Jue Wang, and Qifeng Chen. 2022.
\newblock High-fidelity gan inversion for image attribute editing.
\newblock In \emph{Proceedings of the IEEE/CVF Conference on Computer Vision and Pattern Recognition (CVPR)}.

\bibitem[{Wu et~al.(2023)Wu, Wo, Li, and Han}]{wu2023learning}
Yuanlu Wu, Yan Wo, Caiyu Li, and Guoqiang Han. 2023.
\newblock Learning domain-invariant representation for generalizing face forgery detection.
\newblock \emph{Computers \& Security}, 130:103280.

\bibitem[{Xu et~al.(2025)Xu, Zhang, Li, Tang, Huang, and Zhang}]{xu2024fakeshield}
Zhipei Xu, Xuanyu Zhang, Runyi Li, Zecheng Tang, Qing Huang, and Jian Zhang. 2025.
\newblock Fakeshield: Explainable image forgery detection and localization via multi-modal large language models.
\newblock In \emph{International Conference on Learning Representations}.

\bibitem[{Yan et~al.(2024)Yan, Yao, Chen, Zhao, Fu, Zhu, Luo, Wang, Ding, Wu, and Yuan}]{yan2024df40}
Zhiyuan Yan, Taiping Yao, Shen Chen, Yandan Zhao, Xinghe Fu, Junwei Zhu, Donghao Luo, Chengjie Wang, Shouhong Ding, Yunsheng Wu, and Li~Yuan. 2024.
\newblock \href {https://arxiv.org/abs/2406.13495} {Df40: Toward next-generation deepfake detection}.
\newblock \emph{Preprint}, arXiv:2406.13495.

\bibitem[{Yu et~al.(2018)Yu, Wang, Peng, Gao, Yu, and Sang}]{yu2018bisenet}
Changqian Yu, Jingbo Wang, Chao Peng, Changxin Gao, Gang Yu, and Nong Sang. 2018.
\newblock Bisenet: Bilateral segmentation network for real-time semantic segmentation.
\newblock In \emph{Proceedings of the European conference on computer vision (ECCV)}, pages 325--341.

\bibitem[{Yu et~al.(2021)Yu, Xia, Fei, and Lu}]{yu2021survey}
Peipeng Yu, Zhihua Xia, Jianwei Fei, and Yujiang Lu. 2021.
\newblock A survey on deepfake video detection.
\newblock \emph{Iet Biometrics}, 10(6):607--624.

\bibitem[{Yuan et~al.(2024)Yuan, Li, Liu, Tang, Luo, Qin, Zhang, and Zhu}]{yuan2024osprey}
Yuqian Yuan, Wentong Li, Jian Liu, Dongqi Tang, Xinjie Luo, Chi Qin, Lei Zhang, and Jianke Zhu. 2024.
\newblock Osprey: Pixel understanding with visual instruction tuning.
\newblock In \emph{Proceedings of the IEEE/CVF Conference on Computer Vision and Pattern Recognition}, pages 28202--28211.

\bibitem[{Zeng et~al.(2024)Zeng, Li, Gao, and Pang}]{zeng2024multimodal}
Fengzhu Zeng, Wenqian Li, Wei Gao, and Yan Pang. 2024.
\newblock Multimodal misinformation detection by learning from synthetic data with multimodal llms.
\newblock In \emph{Findings of the Association for Computational Linguistics: EMNLP 2024}, pages 10467--10484.

\bibitem[{Zhang et~al.(2025)Zhang, Gao, Jiang, Zhao, and Zheng}]{zhang2025ctrl}
Guiyu Zhang, Huan-ang Gao, Zijian Jiang, Hao Zhao, and Zhedong Zheng. 2025.
\newblock Ctrl-u: Robust conditional image generation via uncertainty-aware reward modeling.
\newblock \emph{ICLR}.

\bibitem[{Zhang et~al.(2024{\natexlab{a}})Zhang, Yu, Wang, Huang, Shen, Gao, and Ren}]{zhang2024genface}
Yaning Zhang, Zitong Yu, Tianyi Wang, Xiaobin Huang, Linlin Shen, Zan Gao, and Jianfeng Ren. 2024{\natexlab{a}}.
\newblock Genface: A large-scale fine-grained face forgery benchmark and cross appearance-edge learning.
\newblock \emph{IEEE Transactions on Information Forensics and Security}.

\bibitem[{Zhang et~al.(2024{\natexlab{b}})Zhang, Colman, Guo, Shahriyari, and Bharaj}]{zhang2024common}
Yue Zhang, Ben Colman, Xiao Guo, Ali Shahriyari, and Gaurav Bharaj. 2024{\natexlab{b}}.
\newblock Common sense reasoning for deepfake detection.
\newblock In \emph{European Conference on Computer Vision}, pages 399--415. Springer.

\bibitem[{Zhu et~al.(2022)Zhu, Wu, Zhu, Jiang, Tang, Zhang, Liu, and Loy}]{zhu2022celebv}
Hao Zhu, Wayne Wu, Wentao Zhu, Liming Jiang, Siwei Tang, Li~Zhang, Ziwei Liu, and Chen~Change Loy. 2022.
\newblock Celebv-hq: A large-scale video facial attributes dataset.
\newblock In \emph{European conference on computer vision}, pages 650--667. Springer.

\bibitem[{Zhu et~al.(2025{\natexlab{a}})Zhu, Wang, Chen, Liu, Ye, Gu, Tian, Duan, Su, Shao, Gao, Cui, Wang, Cao, Liu, Wei, Zhang, Wang, Xu, Li, Wang, Deng, Li, He, Jiang, Luo, Wang, He, Shi, Zhang, Shao, He, Xiong, Qu, Sun, Jiao, Lv, Wu, Zhang, Deng, Ge, Chen, Wang, Dou, Lu, Zhu, Lu, Lin, Qiao, Dai, and Wang}]{zhu2025internvl3exploringadvancedtraining}
Jinguo Zhu, Weiyun Wang, Zhe Chen, Zhaoyang Liu, Shenglong Ye, Lixin Gu, Hao Tian, Yuchen Duan, Weijie Su, Jie Shao, Zhangwei Gao, Erfei Cui, Xuehui Wang, Yue Cao, Yangzhou Liu, Xingguang Wei, Hongjie Zhang, Haomin Wang, Weiye Xu, and 32 others. 2025{\natexlab{a}}.
\newblock \href {https://arxiv.org/abs/2504.10479} {Internvl3: Exploring advanced training and test-time recipes for open-source multimodal models}.
\newblock \emph{Preprint}, arXiv:2504.10479.

\bibitem[{Zhu et~al.(2025{\natexlab{b}})Zhu, Liu, Zheng, and Liu}]{zhu2025seed}
Yule Zhu, Ping Liu, Zhedong Zheng, and Wei Liu. 2025{\natexlab{b}}.
\newblock Seed: A benchmark dataset for sequential facial attribute editing with diffusion models.
\newblock \emph{arXiv preprint arXiv:2506.00562}.

\end{thebibliography}

\appendix

\clearpage
\setcounter{page}{1}

\pagestyle{plain}


\section{Related Work}
\label{Related Work}

\noindent\textbf{Facial Manipulation Localization.} Detecting manipulated facial regions, especially deepfakes, has garnered attention. CNN-based methods \citep{sabir2019recurrent} utilize temporal inconsistencies for videos, while GAN-based approaches, such as GANprintR \citep{neves2020ganprintr} and MaskGAN \citep{liu2022maskgan}, address synthetic artifacts. Hybrid models like HCiT \citep{kaddar2021hcit} combine CNNs and ViTs to enhance generalization, and multi-modal methods \citep{sun2023face, khedkar2022exploiting} leverage spatial-temporal inconsistencies. However, these models lack interpretability and fine-grained mask generation, which our work addresses by providing both localization masks and textual explanations.

\noindent\textbf{Multi-label Classification for Facial Localization.} Multi-label classification captures independent alterations in facial regions but struggles with dependencies across features. CNNs \citep{lalitha2022deepfake} face limitations in fine-grained tasks, while hybrid models \citep{kaddar2021hcit} improve detection by combining local and global features. Weighted loss functions \citep{ramachandran2021experimental} and parallel branches \citep{richards2023deep} address class imbalance and refine detection. Yet, few works integrate multi-label classification with localization. Our ViT-based classifier bridges this gap by capturing complex dependencies with parallel branches and weighted loss functions.

\noindent\textbf{Segmentation Techniques.} Segmentation is crucial for identifying localized manipulations. Models like U-Net and DeepLab \citep{ross2017focal} focus on spatial features, while Transformer models \citep{alexey2020image} capture global context. Recent methods like SAM \citep{kirillov2023segment} use a Two-Way Transformer for high-quality masks but lack manipulation-specific context. By integrating SAM with InstructBLIP, we create context-aware forgery masks, unifying segmentation and manipulation detection for enhanced localization.

\noindent\textbf{Explainable Forgery Detection.} A recent trend is moving towards explainable forensics. Within the NLP community, multimodal misinformation detection aims to identify inconsistencies between text and images via logic reasoning~\citep{liu2023interpretable} or synthetic data learning~\citep{zeng2024multimodal}, yet these methods often overlook pixel-level manipulation artifacts. In the vision domain, FakeShield~\citep{xu2024fakeshield} attempts to explain general image forgeries but relies on synthetic GPT-4o annotations. Bridging these fields, our work introduces \textbf{MMTT}, a large-scale dataset specifically designed for the facial forgery domain. Unlike previous resources, MMTT features meticulous human-in-the-loop annotation, enabling models to ground their textual explanations in precise visual regions.

\section{Forgery-aware Pretraining}

In the main paper, we briefly outlined the Forgery-aware Pretraining stage. To provide a more comprehensive understanding of our optimization strategy, this section details the specific mathematical formulations for the four distinct training objectives: Masked Language Modeling, Language Modeling, Forgery Localization, and Cross-model Alignment Learning.


The goal of our forgery-aware pretraining stage is to learn robust multimodal representations that are sensitive to manipulation artifacts. Given an image \(I\) and its corresponding ground-truth explanation text \(T\), we jointly optimize the core modules of our model using four distinct training objectives. The image \(I\) is first processed by a frozen visual encoder to yield embeddings \(E_I\), which serve as input alongside the text \(T\) for the following loss functions:

 \textbf{Masked Language Modeling (\(\mathcal{L}_{mlm}\)):} 
    The text \(T\) is tokenized into \(\tilde{T}\). Before feeding \(\tilde{T}\) into the Q-Former, a subset of region-related tokens (\eg, “ear”, “eye”, etc.)  \(\mathcal{M}\)  is masked, and the masked token results in \(\tilde{T}_{\setminus \mathcal{M}}\). Along with the learned query tokens \(Q\) and image embeddings \(E_I\), the Q-Former predicts the masked tokens. The loss is computed as:
    \begin{equation}
    \label{eq:mlm_loss}
        \mathcal{L}_{mlm} = -\sum_{t \in \mathcal{M}} \log P(t \mid I, \tilde{T}_{\setminus \mathcal{M}}).
    \end{equation}
\textbf{Language Modeling (\(\mathcal{L}_{lm}\)):} 
    The Q-Former output is projected and fed to a T5-based decoder~\citep{chung2022scaling} that generates the {explanatory text} \(\hat{T}\) {with the length of \(L_{\hat{T}}\)}. The generated explanation is compared token-by-token with the ground truth via cross-entropy loss:
    \begin{equation}
    \label{eq:lm_loss_single}
    \mathcal{L}_{lm} = -\sum_{k=1}^{L_{\hat{T}}} \log P\Big(\hat{T}_k \mid I, \hat{T}_0, \ldots, \hat{T}_{k-1}\Big),
    \end{equation}
\textbf{Forgery Localization (\(\mathcal{L}_{seg}\)):} 
    The non-[CLS] tokens of \(E_I\) are seamlessly fused with the text $T$ via cross-attention. The mask decoder predicts a forgery mask \(\hat{M}\) {with the height \(H\) and width \(W\)}, which is compared to the ground-truth mask \(M\) using pixel-wise cross-entropy loss:
    \begin{equation}
    \scriptsize
    \label{eq:seg_loss}
    \mathcal{L}_{seg} = -\frac{1}{HW}\sum_{i=1}^{H}\sum_{j=1}^{W}\Big[M_{ij}\log \hat{M}_{ij} + (1-M_{ij})\log(1-\hat{M}_{ij})\Big],
    \end{equation}
    where 
    $M_{ij}=1$ if the $(i,j)$ pixel is manipulated, 0 otherwise.
\textbf{Cross-model Alignment Learning (\(\mathcal{L}_{con}\)):} 
    To align modalities, we pull the global image feature \(v\) (from the [CLS] token) closer to the mean-pooled text feature \(t\) with contrastive loss as: 
    \begin{equation}
    \label{eq:con_loss_batch}
    \mathcal{L}_{con} = -\frac{1}{N} \sum_{i=1}^{N} \log \frac{\exp\Big(\text{sim}(v_i, t_i)/\tau\Big)}{\sum_{j=1}^{N} \exp\Big(\text{sim}(v_i, t_j)/\tau\Big)},
    \end{equation}
    where {$N$ is the batch size, } \(\text{sim}(\cdot)\) {denotes} cosine similarity and \(\tau\) is a temperature parameter.
The overall pretraining loss is defined as:
\begin{equation}
\label{eq:pretrain_loss}
\mathcal{L}_{pretrain} = \lambda_{1}\mathcal{L}_{mlm} + \lambda_{2}\mathcal{L}_{lm} + \lambda_{3}\mathcal{L}_{seg} + \lambda_{4}\mathcal{L}_{con},
\end{equation}
where \(\lambda_{1}\), \(\lambda_{2}\), \(\lambda_{3}\), and \(\lambda_{4}\) being empirically tuned weights.
The joint optimization of these losses enables our model to capture both fine-grained local details and global semantic context. This robust initialization is pivotal for the subsequent {Attribution Report Generation Stage}, where further fine-tuning refines forgery localization and enhances the quality of the generated {reports}.

\section{Examples from MMTT Dataset}

\begin{figure*}[ht]
    \centering
    \includegraphics[width=1.0\linewidth]{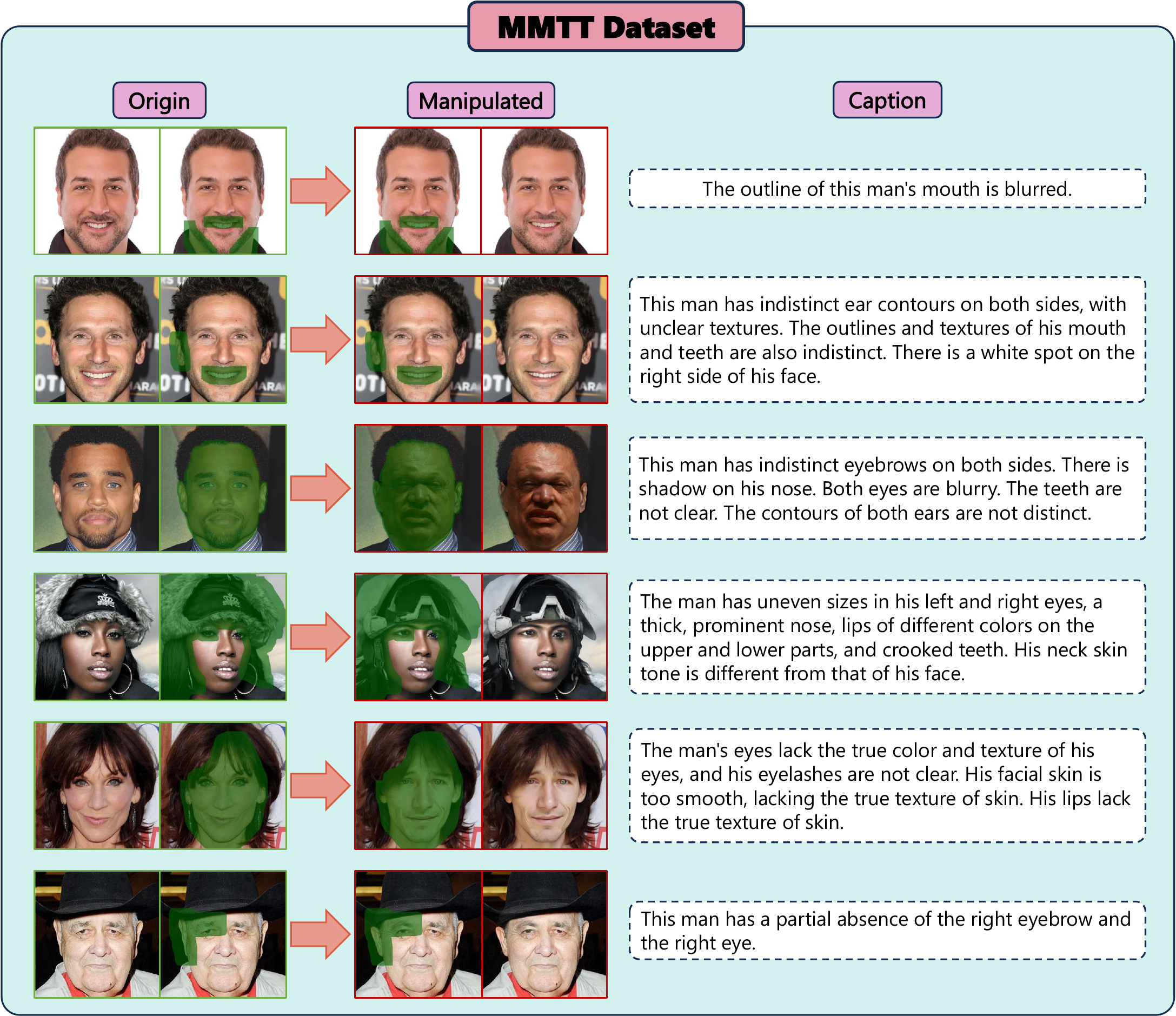}
    \caption{Examples from the MMTT dataset. Each row illustrates a case from the dataset, comprising a manipulated image, its corresponding binary mask (overlaid in green), and a textual description detailing the altered facial regions. For illustrative purposes, the original (authentic) images are also included in this figure to highlight the extent and nature of the manipulations. The green regions indicate the localized areas of forgery as identified by the binary masks. It is important to note that the original images are not part of the MMTT dataset; the dataset itself consists only of manipulated images, binary masks, and their associated textual descriptions.}
    \label{fig:more_dataset}
\end{figure*}


To enhance the understanding of the MMTT dataset and its unique contributions to facial image forgery localization, we provide a word cloud generated from the textual descriptions (captions) and a series of representative examples. The MMTT dataset is meticulously designed to facilitate fine-grained forgery localization by leveraging multimodal annotations. Each sample consists of three complementary components: a manipulated image, a binary mask delineating the forged regions, and a detailed textual description that explicitly identifies and contextualizes the alterations. These comprehensive annotations provide a robust foundation for research tasks requiring precise localization and explainability of facial manipulations.

\begin{figure*}[t] 
    \centering
    
    \includegraphics[width=\textwidth]{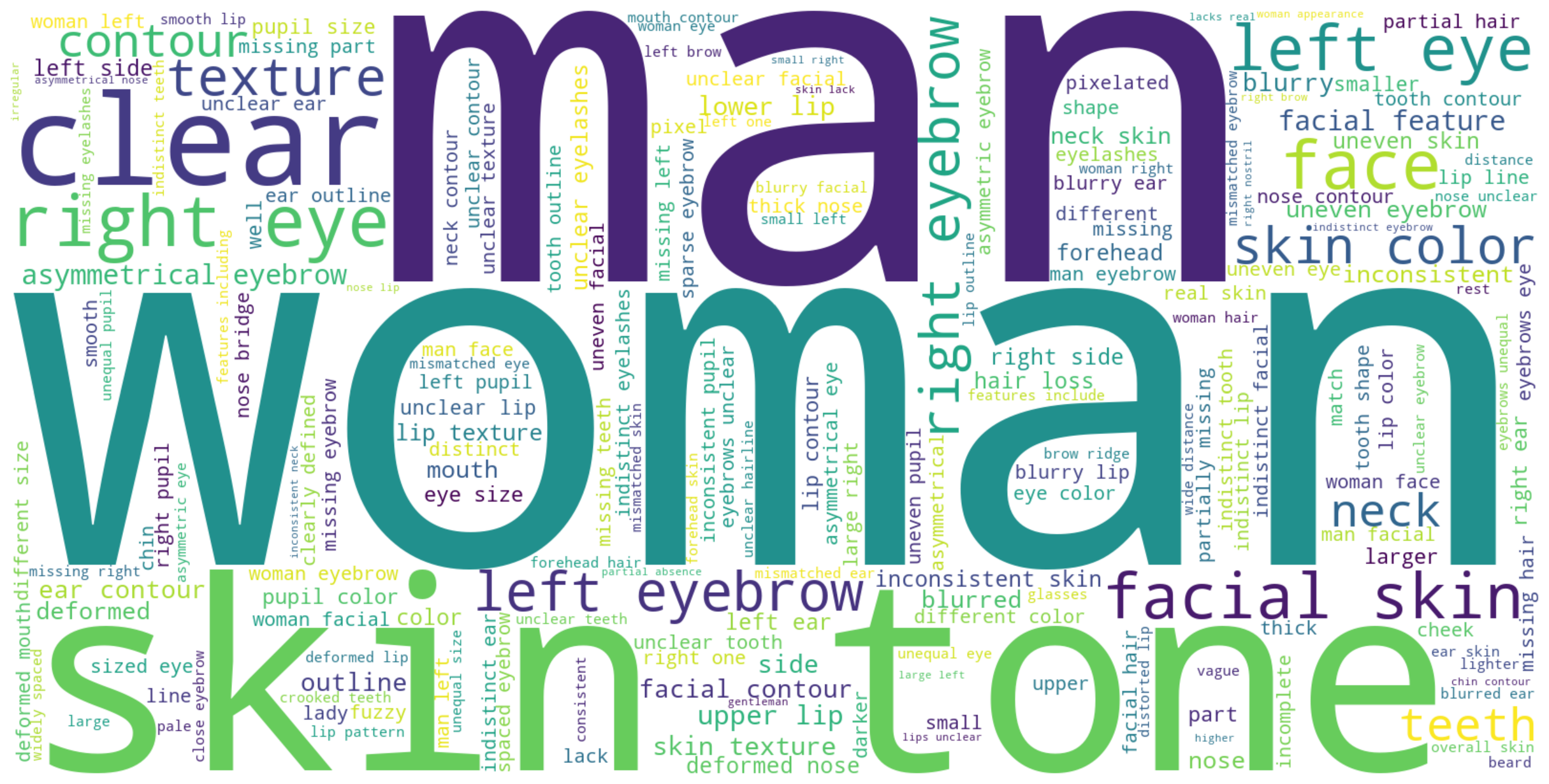}
    
    \caption{Word cloud of captions in the MMTT dataset. This visualization highlights the most frequently used words in the dataset's textual descriptions, where the font size represents the frequency of occurrence.}
    \label{fig:wordcloud}
\end{figure*}

The word cloud, presented in Figure~\ref{fig:wordcloud}, visually encapsulates the linguistic distribution within the dataset’s textual annotations. Dominant terms such as "woman," "man," "skin tone," and "facial skin" highlight the dataset’s focus on describing forgery in specific facial regions. Furthermore, frequent mentions of region-specific features, such as "left eye," "nose bridge," and "right eyebrow," underscore the granularity and specificity of the annotations. This visualization demonstrates the alignment between the textual descriptions and the underlying task of forgery localization, offering an overview of the dataset’s descriptive richness and consistency.

Figure~\ref{fig:more_dataset} illustrates selected examples from the MMTT dataset, showcasing its multimodal structure and the diversity of forgery types. Each example includes a manipulated image, its corresponding binary mask, and a textual description. For illustrative purposes, we have also included the original (authentic) images alongside the manipulated samples in Figure~\ref{fig:more_dataset} to provide additional context for understanding the nature and extent of the forgeries. It is important to note that these original images are not part of the MMTT dataset and are shown exclusively to highlight the transformations and to provide clarity on the dataset’s structure. The actual dataset is focused on forged images, binary masks, and detailed captions, without the inclusion of original (authentic) images.

\section{Experimental Setup}

\noindent\textbf{Implementation Details.} We implement ForgeryTalker with PyTorch \citep{pytorch} and train on a single NVIDIA H100 (94GB) GPU. Our model is built upon the InstructBLIP framework, utilizing the frozen Flan-T5-XL as the Large Language Model (LLM) backbone. The total number of parameters is approximately 4B. The entire two-stage training process takes approximately 2 days (i.e., 48 GPU hours).

\noindent\textbf{Training Protocol.} We use an 8:1:1 train/validation/test split of the MMTT dataset. The two-stage training process is as follows: (1) The FPN is trained for 125k steps (batch size 16, initial lr 7.5e-3 with cosine decay) with the BCE loss weight $\omega$ set to 0.2. (2) With the FPN frozen, the main model is trained for 60 epochs (batch size 16, lr 4e-6) using mixed-precision (fp16) training.

\noindent\textbf{Evaluation Metrics.} For the report generation task, we employ standard captioning metrics including CIDEr, BLEU, and ROUGE-L, computed using the official \texttt{pycocoevalcap} toolkit. For forgery localization, we report pixel-level Intersection over Union (IoU), Precision, and Recall to evaluate the mask alignment quality.

\section{Comparison with State-of-the-art Models}

We compare the performance of our proposed ForgeryTalker framework against a range of recent models, including specialist forgery localization models and general-purpose Large Vision-Language Models (LVLMs). It is important to note that while the specialist models and our ForgeryTalker were trained on our dataset, the general LVLMs were evaluated in a zero-shot setting to assess their out-of-the-box capabilities for this novel task. The evaluation covers both forgery localization (IoU, Precision, Recall) and interpretation generation (CIDEr, ROUGE-L, BLEU scores), with results summarized in Table~\ref{tab:ComparisonSupplementary}.

For forgery localization, specialist models like IML-ViT expectedly achieve the highest scores in IoU (77.89) and Recall (90.04), as they are solely optimized for this task. However, our ForgeryTalker demonstrates highly competitive localization capabilities, achieving a strong IoU of \textbf{73.67} and securing the best Precision score (\textbf{91.43}) among all compared models. This indicates our model's superior ability to avoid over-predicting forged regions.

For the primary task of interpretation generation, ForgeryTalker significantly outperforms all other LVLMs across every text-based metric. It achieves a CIDEr score of \textbf{59.3}, which is an order of magnitude higher than the next best competitor, Llava-72B (3.06). This substantial gap highlights the effectiveness of our forgery-aware architecture in generating accurate and relevant textual explanations, a task where general-purpose LVLMs, which were not fine-tuned on our forgery-specific data, naturally struggle. In summary, ForgeryTalker establishes a new state-of-the-art in interpretable forgery localization by providing best-in-class captioning performance while maintaining a robust and precise localization ability.

\begin{table*}[t]
\centering
\resizebox{\textwidth}{!}{%
\begin{tabular}{l|cccccc|ccc}
\toprule[1.5pt]
\multirow{2}{*}{\textbf{Model}} & \multicolumn{6}{c|}{\textbf{Interpretation Generation}} & \multicolumn{3}{c}{\textbf{Forgery Localization}} \\
\cmidrule(r){2-7} \cmidrule(r){8-10}
& {CIDEr} & {ROUGE\_L} & {Bleu\_1} & {Bleu\_2} & {Bleu\_3} & {Bleu\_4} & {IoU} & {Precision} & {Recall} \\
\hline
Seed1.5VL~\citep{guo2025seed15vltechnicalreport} & 0.54 & 13.74 & 17.19 & 5.35 & 1.86 & 0.98 & - & - & - \\
Qwen2.5VL (72B)~\citep{bai2025qwen25vltechnicalreport} & 2.72 & 16.52 & 20.34 & 6.33 & 2.55 & 1.46 & - & - & - \\
Qwen2.5VL (32B)~\citep{bai2025qwen25vltechnicalreport} & 2.53 & 16.89 & 22.44 & 8.16 & 3.3 & 1.78 & - & - & - \\
Qwen2.5VL (7B)~\citep{bai2025qwen25vltechnicalreport} & 2.48 & 17.0 & 22.33 & 8.2 & 3.24 & 1.78 & - & - & - \\
llava (72B)~\citep{liu2024llavanext} & 3.06 & 16.83 & 22.01 & 8.35 & 3.3 & 1.8 & - & - & - \\
llava (8B)~\citep{liu2024llavanext} & 2.1 & 18.05 & 20.75 & 7.66 & 3.15 & 1.75 & - & - & - \\
InternVL3 (78B)~\citep{zhu2025internvl3exploringadvancedtraining} & 2.67 & 16.82 & 22.31 & 8.08 & 3.16 & 1.75 & - & - & - \\
InternVL3 (38B)~\citep{zhu2025internvl3exploringadvancedtraining} & 2.29 & 17.18 & 21.21 & 7.83 & 3.2 & 1.76 & - & - & - \\
InternVL3 (14B)~\citep{zhu2025internvl3exploringadvancedtraining} & 2.29 & 16.85 & 20.81 & 7.54 & 3.03 & 1.71 & - & - & - \\
IML-ViT~\citep{ma2023imlvit} & - & - & - & - & - & - & \textbf{77.89} & 83.76 & \textbf{90.04} \\
PSCC-NET~\citep{liu2022pscc} & - & - & - & - & - & - & 32.33 & 70.3 & 37.44 \\
\hline
\textbf{ForgeryTalker} & \textbf{59.3} & \textbf{28.8} & \textbf{35.0} & \textbf{22.1} & \textbf{16.0} & \textbf{12.5} & 73.67 & \textbf{91.43} & 86.22 \\
\bottomrule[1.5pt]
\end{tabular}}
\caption{Performance comparison against state-of-the-art models in the supplementary material. Specialist models (IML-ViT, PSCC-NET) and our ForgeryTalker are trained on our dataset. \textbf{General Large Vision-Language Models (LVLMs) are evaluated in a zero-shot setting.} Best performance for each metric is in \textbf{bold}.}
\label{tab:ComparisonSupplementary}
\end{table*}



\section{Ablation Study on the Impact of the FPN}

\noindent\textbf{Impact of FPN Loss and Discount Factor.}  
We use Positive Label Matching (PLM) to evaluate the effectiveness of FPN. PLM calculates the ratio of correctly predicted positive labels over the union of predicted and ground-truth positive labels:

\vspace{-0.2cm}
\begin{equation}
\scriptsize
\label{PLM}
    \text{PLM} = \frac{|\text{Predicted Positive Labels} \cap \text{Ground Truth Positive Labels}|}{|\text{Predicted Positive Labels} \cup \text{Ground Truth Positive Labels}|}.
\end{equation}

\begin{table}[h!] 
    \centering
    \resizebox{\linewidth}{!}{
        \begin{tabular}{cccc}
        \toprule[1.5pt]
        \textbf{Model} & \boldmath{$\omega$} & \textbf{Loss} & \textbf{PLM} \\
        \hline
        ViT & 1 & BCE & 34.23 \\
        ViT & 0.2 & BCE & 38.92 \\
        FPN & 0.2 & BCE & 39.16 \\
        \hline
        \textbf{FPN} & 0.2 & BCE + Dice & \textbf{41.05} \\
        \bottomrule[1.5pt]
        \end{tabular}
    }
    \caption{Ablation Study on the Impact of the FPN}
    \label{tab:AblationStudyFPN}
\end{table}

Unlike IoU, PLM focuses on detecting manipulated regions without being influenced by a large number of correctly predicted negative labels, making it ideal for tasks with sparse modifications.

The forgery prompter network is optimized by a combined loss, incorporating both Binary Cross-Entropy (BCE) loss and Dice loss to effectively balance region classification and overlap precision:
\begin{equation}
\label{bceloss_2}
    \mathcal{L}_{BCE} = -\frac{1}{21}\sum_{i=1}^{21} Y_i\log \hat{Y}_i + \omega (1-Y_i)\log (1-\hat{Y}_i),
\end{equation}
where $\omega$ is a discount factor set to $\omega < 1$ to address the imbalance due to the prevalence of unmodified regions.

\begin{table*}[t!]
\centering
\footnotesize
\begin{tabular}{lcccccc}
\toprule[1.5pt] 
\textbf{Model} & \textbf{Bleu\_1} & \textbf{Bleu\_2} & \textbf{Bleu\_3} & \textbf{Bleu\_4} & \textbf{ROUGE\_L} & \textbf{CIDEr} \\
\hline
SCA~\citep{huang2024segment} & 9.23 & 6.45 & 3.89 & 0.93 & 15.74 & 0.54 \\
Osprey~\citep{yuan2024osprey} & 7.54 & 6.09 & 4.36 & 2.29 & 13.28 & 1.40 \\ \hline
\textbf{ForgeryTalker} & \textbf{10.80} & \textbf{8.80} & \textbf{7.40} & \textbf{6.50} & \textbf{32.20} & \textbf{2.20} \\
\bottomrule[1.5pt]
\end{tabular}
\caption{Zero-shot report generation performance on the face-modification subset of the SynthScars dataset~\cite{kang2025legionlearninggroundexplain}. All models were fine-tuned on the MMTT dataset and evaluated on SynthScars without further tuning.}
\label{tab:synthscars}
\end{table*}

\begin{table*}[h!]
\centering
\footnotesize
\caption{Ablation study on the contribution of different forgery types within the MMTT dataset. Each row represents a model trained on a specific subset of data (leaving one type out) and evaluated on the full test set. ``Full Set'' indicates training with all forgery types.}
\label{tab:mmtt_contribution}
\resizebox{\textwidth}{!}{%
\begin{tabular}{l|ccc|cccccc}
\toprule[1.5pt]
\multirow{2}{*}{\textbf{Training Data Composition}} & \multicolumn{3}{c|}{\textbf{Forgery Localization}} & \multicolumn{6}{c}{\textbf{Report Generation}} \\
\cmidrule(l){2-4} \cmidrule(l){5-10}
& \textbf{IoU} & \textbf{Precision} & \textbf{Recall} & \textbf{Bleu\_1} & \textbf{Bleu\_2} & \textbf{Bleu\_3} & \textbf{Bleu\_4} & \textbf{ROUGE\_L} & \textbf{CIDEr} \\
\hline
\textbf{Full Set (All Types)} & \textbf{71.00} & \textbf{90.27} & \textbf{84.04} & \textbf{31.90} & \textbf{19.60} & \textbf{14.20} & \textbf{11.20} & \textbf{28.30} & \textbf{56.90} \\
\hline
w/o Face Swapping & 63.12 & 84.81 & 73.80 & 28.10 & 17.80 & 12.70 & 10.00 & 26.30 & 49.20 \\
w/o Image Inpainting & 61.28 & 72.95 & 80.55 & 25.00 & 15.90 & 11.80 & 9.50 & 2.70 & 48.50 \\
w/o Face Editing & 68.20 & 89.77 & 78.72 & 23.50 & 12.30 & 7.00 & 4.20 & 21.50 & 16.20 \\
\bottomrule[1.5pt]
\end{tabular}%
}
\end{table*}

Table~\ref{tab:AblationStudyFPN} examines the effect of the discount factor $\omega$ in the BCE loss (Eq.~\ref{bceloss_2}) and the addition of Dice loss. Setting $\omega=0.2$ improves the PLM metric from 34.23 to 38.92 on a ViT backbone; further incorporating the FPN boosts PLM to 39.16, and combining BCE with Dice raises it to 41.05. This confirms that discounting unmodified regions and combining losses enhances region prompt accuracy.

\section{Evaluation on SynthScars Dataset}

To further verify the robustness of our method, we extended our evaluation to the SynthScars dataset~\cite{kang2025legionlearninggroundexplain} in a zero-shot setting, specifically filtering for samples involving facial manipulations. We benchmark ForgeryTalker against SCA and Osprey. For a fair comparison, both baselines were also fine-tuned on our MMTT dataset before evaluation. As reported in Table~\ref{tab:synthscars}, ForgeryTalker consistently outperforms the baselines across all report generation metrics. Specifically, it achieves a ROUGE-L of 32.2 and BLEU-4 of 6.5, significantly surpassing the second-best results. It is worth noting that the absolute scores for n-gram based metrics, particularly CIDEr (2.2), are relatively modest for all models. This is largely attributed to the inherent stylistic discrepancy between the ground-truth captions in SynthScars and our MMTT training data. Since metrics like CIDEr are highly sensitive to specific linguistic patterns, the domain shift in annotation style leads to lower numerical scores. Despite this, our method demonstrates superior transferability and relative performance compared to the state-of-the-art baselines.

\section{Ablation Study on Manipulation Type Contributions}

To investigate the individual contributions of different forgery types within our MMTT dataset—namely Face Swapping, Face Editing, and Image Inpainting—we conducted an ablation study based on data composition. specifically, we trained ForgeryTalker on subsets containing only two out of the three manipulation types and evaluated the performance on the full test set. This ``leave-one-type-out'' setting allows us to quantify the impact of the missing type on the model's generalization capability.

The results are presented in Table~\ref{tab:mmtt_contribution}. The comparison reveals distinct roles for each manipulation type:
\begin{itemize}
    \item \textbf{Impact of Face Editing on Report Quality:} When Face Editing data is excluded from training (see the column ``Swapping + Inpainting''), the report generation performance suffers the most dramatic decline, with the CIDEr score dropping from 56.9 to \textbf{16.2}. This suggests that the Face Editing samples in MMTT contain the most diverse and semantically complex linguistic descriptions. Without them, the model struggles to generate high-quality, descriptive attribution reports.
    \item \textbf{Impact of Inpainting on Localization and Fluency:} Excluding Image Inpainting data (see the column ``Swapping + Editing'') leads to the lowest localization accuracy (IoU drops to 61.28) and a collapse in ROUGE-L score (2.7). This indicates that Inpainting samples are critical for the model to learn precise boundary localization and to maintain the structural fluency of the generated text.
    \item \textbf{Impact of Face Swapping:} Removing Face Swapping data (see the column ``Editing + Inpainting'') results in a moderate performance drop across both localization and generation metrics (e.g., IoU drops to 63.12, CIDEr to 49.2). This implies that Face Swapping contributes to the overall robustness of the model but is less specialized than the other two types in driving specific extreme metrics.
\end{itemize}
In summary, the diversity of manipulation types in MMTT is essential, with each type contributing uniquely to either the visual localization precision or the semantic richness of the generated reports.

\section{Statements}

\subsection{Reproducibility Statement}
To ensure transparency and facilitate future research, we commit to the full release of our code, data, and model weights. First, we will make the complete source code for ForgeryTalker publicly available on GitHub upon publication, covering both the pre-training and report generation stages. This repository will include comprehensive training scripts, configuration files for ablation studies, and detailed environment setup instructions. Second, the MMTT dataset, which consists of 152,217 image-text-mask triplets, will be released to the research community under our Data Usage Agreement. Finally, to establish a standardized benchmark and lower the barrier for subsequent work, we will provide pre-trained checkpoints for our best-performing models and baselines.

\subsection{LLM Usage Statement}

During the preparation of this work, the authors used a large language model to assist with improving grammar, rephrasing sentences, and ensuring terminological consistency. The authors reviewed and edited all model-generated text and take full responsibility for the final content of this paper.

\end{document}